\documentclass[12pt]{article}
\let\OLDthebibliography\thebibliography
\renewcommand\thebibliography[1]{
    \OLDthebibliography{#1}
    \setlength{\parskip}{0pt}
    \setlength{\itemsep}{0pt plus 0.3ex}
}

\usepackage{multirow}
\usepackage[margin=0.9in]{geometry}
\usepackage[utf8]{inputenc}
\usepackage{booktabs}
\usepackage{caption}
\usepackage{subcaption}
\usepackage{adjustbox}
\usepackage{lipsum}
\usepackage[parfill]{parskip}
\usepackage{pgfplots}
\usepackage[labelfont=bf]{caption}

\title{\large \textbf{Integrated Optimization of Predictive and Prescriptive Tasks}\thanks{\href{https://www.abstractsonline.com/pp8/\#!/6818/presentation/11528}{\color{red}{This work was presented in INFORMS Annual Meeting, Seattle, Washington, October 20-23, 2019.}}}}

\author{Mehmet Kolcu\\ 
{\small Industrial and Systems Engineering, Wayne State University, Detroit, MI 48201}\\
{\small mehmet.kolcu@wayne.edu}\\
\and 
Alper E. Murat\\ 
{\small Industrial and Systems Engineering, Wayne State University, Detroit, MI 48201}\\
{\small amurat@wayne.edu}\\
}
\usepackage{graphicx}
\graphicspath{ {./images/} }
\usepackage{amsmath}
\usepackage[bookmarks]{hyperref}
\usetikzlibrary{positioning,fit,arrows.meta,backgrounds}
\usepackage[linesnumbered,ruled,vlined]{algorithm2e}
\DeclareMathOperator*{\argmin}{argmin}
\usepackage{setspace} 
\usepackage{amsfonts}
\pgfplotsset{compat=1.16} 
\tikzset{
		module/.style={%
		draw, rounded corners,
        minimum width=#1,
        minimum height=7mm,
        font=\sffamily
        },
    module/.default=2cm,
    >=LaTeX
}
\begin{document}
\maketitle
\doublespacing
\begin{abstract}
\noindent In traditional machine learning techniques, the degree of closeness between true and predicted values generally measures the quality of predictions.  However, these learning algorithms do not consider prescription problems where the predicted values will be used as input to decision problems. In this paper, we efficiently leverage feature variables, and we propose a new framework directly integrating predictive tasks under prescriptive tasks in order to prescribe consistent decisions. We train the parameters of predictive algorithm within a prescription problem via bilevel optimization techniques.  We present the structure of our method and demonstrate its performance using synthetic data compared to classical methods like point-estimate-based, stochastic optimization and recently developed machine learning based optimization methods. In addition, we control generalization error using different penalty approaches, and optimize the integration over validation data set.
\end{abstract}

\section{Introduction}
Today we are living in a world called the information age. The exponential growth of data availability, ease of accessibility in computational power, and more efficient optimization techniques have paved the way for massive developments in the field of predictive analytic. Particularly when organizations have realized the benefits of predictive methods in improving their efficiency and gaining an advantage over their competitors, these predictive techniques become more powerful. There is a mutual relationship between predictive and prescriptive analytics. We cannot deny the role of optimization techniques while obtaining predictive models because most of the predictive models are trained over the minimization of a loss or maximization of a gain function. On the other hand, prescriptive models containing uncertainty need the estimated inputs from predictive analytics to handle uncertainty in optimization parameters. Although these statistical (machine) learning methods have been provided well-aimed predictions for uncertain parameters in many different fields of science, scenario-based stochastic optimization introduced by \cite{RePEc:inm:ormnsc:v:1:y:1955:i:3-4:p:197-206} and similar works in \cite{10.5555/2031490}, \cite{SHAPIRO2003353}, \cite{doi:10.1137/S1052623499363220}, and \cite{Shapiro2005} are widely preferred techniques in order to tackle uncertainty in decision problems. One of the reasons why these stochastic or robust optimization methods or similar solutions provided by \cite{BEN:09} and \cite{doi:10.1137/080734510} do perform better than point estimate-based optimization is that training process of statistical learning methods does not take into account optimal actions because traditional learning algorithms measure prediction quality based on the degree of closeness between true and predicted values. This gap between prescriptive and predictive analytics leads point estimate-based decisions to a failure in the prescription phase.

There are different variables and parameters in prediction and prescription tasks. Prediction tasks mainly have components such as feature data, response data, and the parameters connecting these two via a function. Prescription problems generally have cost vector parameters in the objective function, right-hand-side and left-hand-side parameters in the constraints, and decision variables. Our interest in this paper is to fit a predictive function with responses and feature data, and feed prescriptive model parameters by this fitted function.

For the sake of uniformity, we use the same notation the rest of this paper. We indicated feature data with $X$, responses with $Y$, parameter of predictive algorithm with $\beta$, and decision variables with $Z$. Responses ($Y$) are the bridge parameter connecting predictive and prescriptive tasks.

\begin{figure}
\makebox[\linewidth][c]{%
\begin{subfigure}[b]{.42\textwidth}
\centering
\includegraphics[width=.95\textwidth]{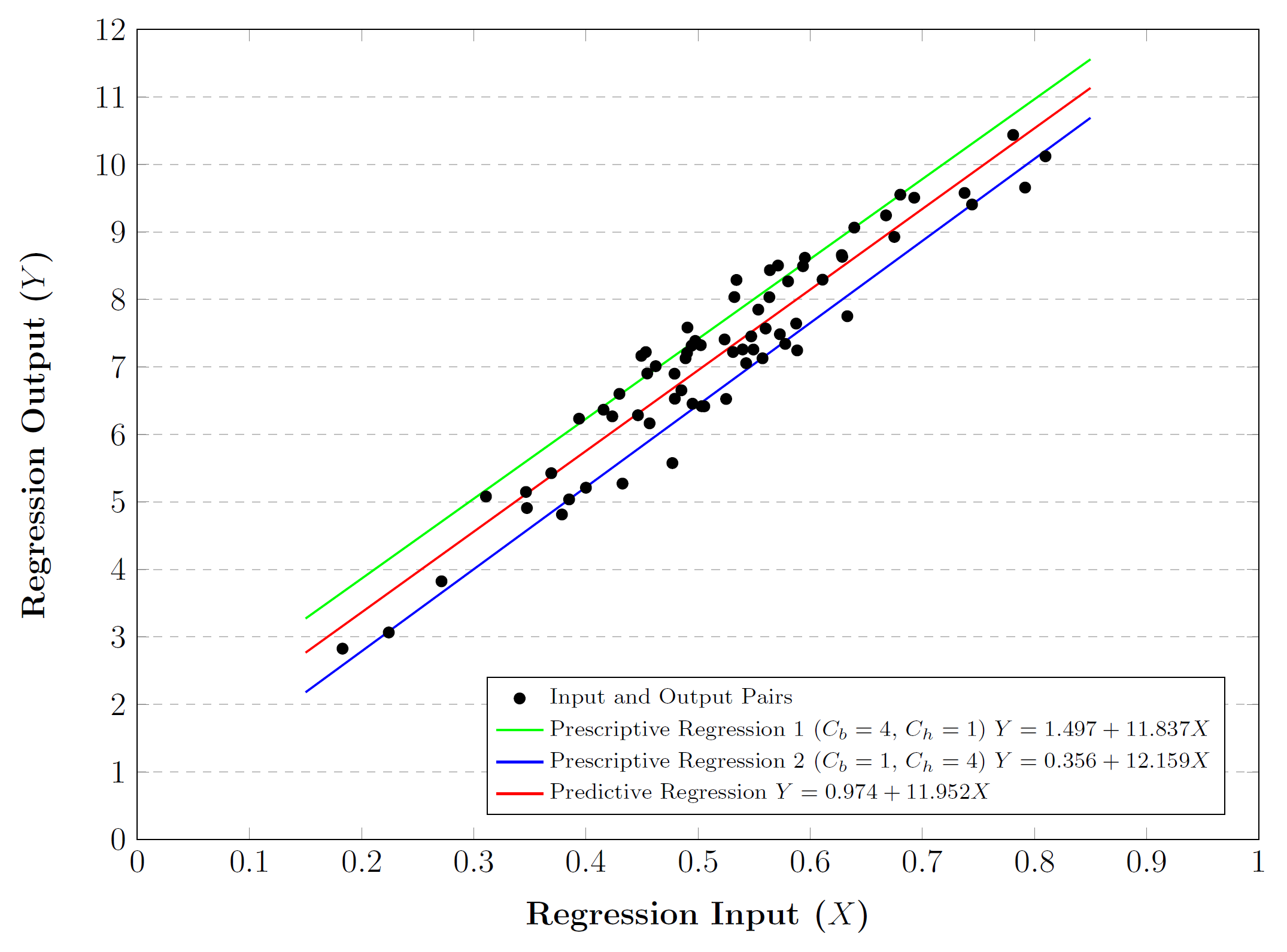}
\caption{Regression Models}
\label{plot:graph11}
\end{subfigure} \hspace{-5mm}%
\begin{subfigure}[b]{.42\textwidth}
\centering
\includegraphics[width=.95\textwidth]{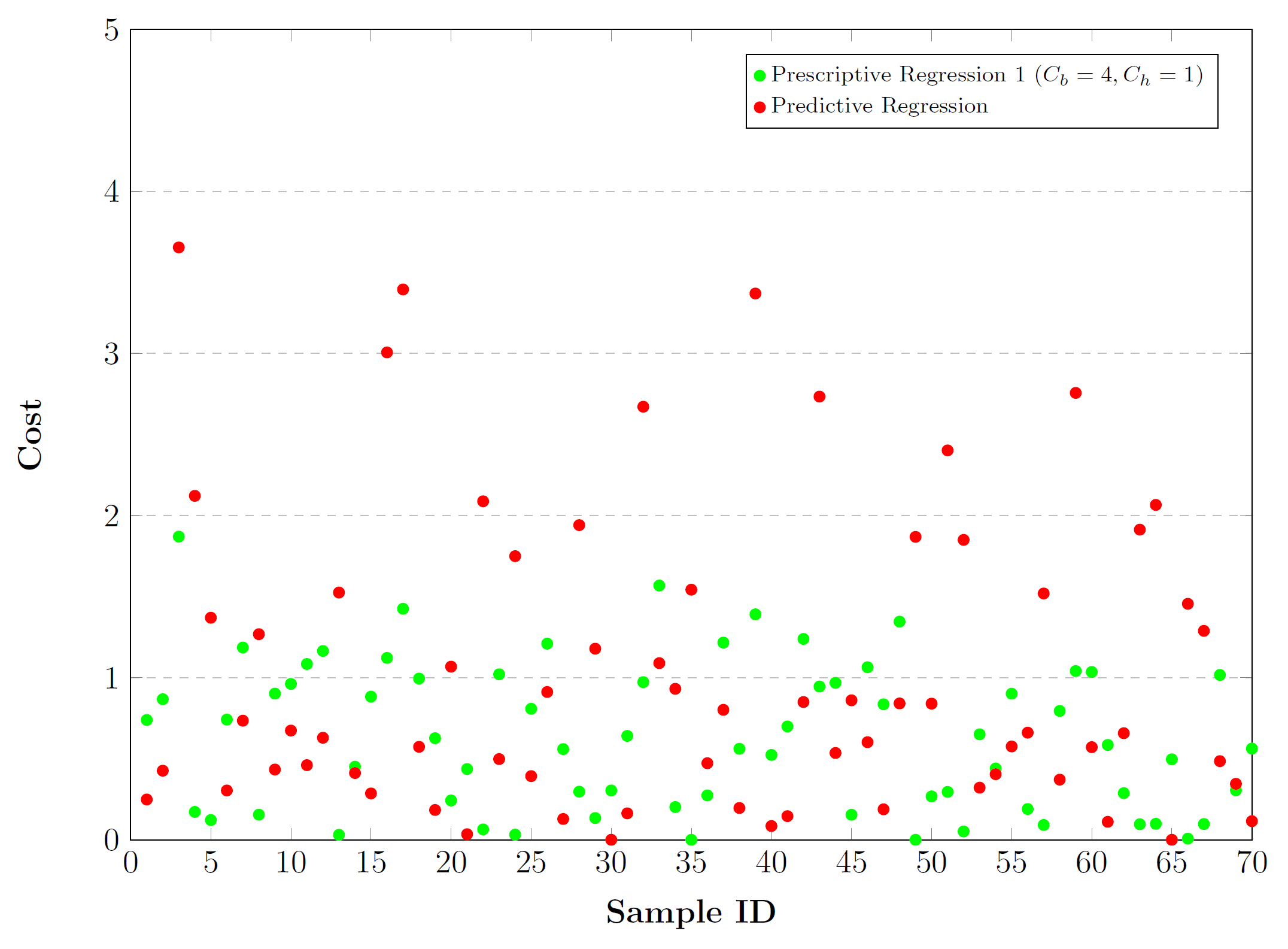}
\caption{Predictive Reg. vs Prescriptive Reg 1 }
\label{plot:graph12}
\end{subfigure} \hspace{-5mm}%
\begin{subfigure}[b]{.42\textwidth}
\centering
\includegraphics[width=.95\textwidth]{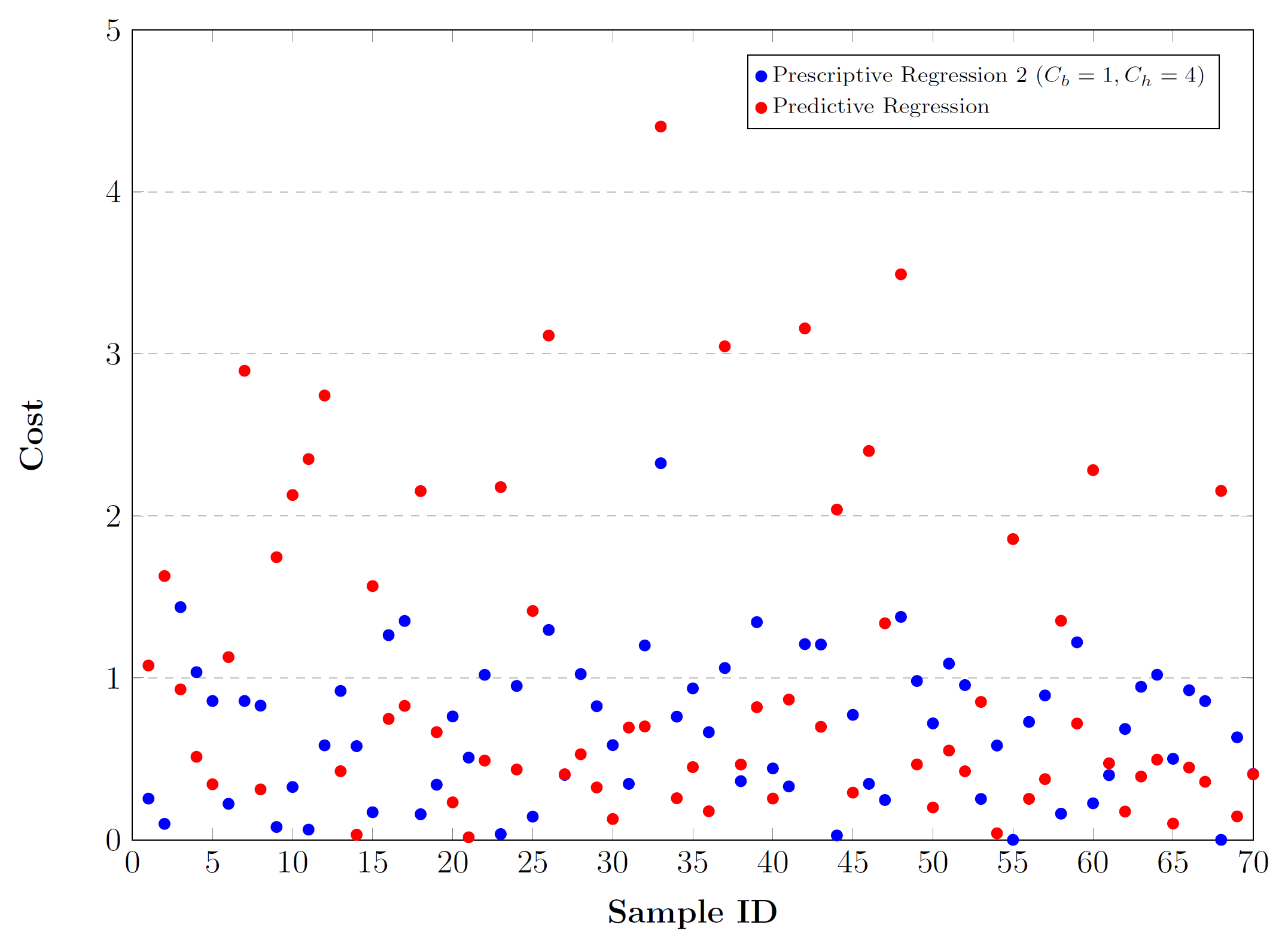}
\caption{Predictive Reg. vs Prescriptive Reg 2 }
\label{plot:graph13}
\end{subfigure}%
}\\
\caption{Comparison of Predictive and Prescriptive Regression Models Based on Cost Function}
\label{plot:graph1}
\end{figure}

In order to visualize our motivation, imagine a newsvendor problem with a cost function, $Cost=C_h(Z-Y)^++C_b(Y-Z)^+$, containing holding cost ($C_h$) and backordering cost ($C_b$) with demand parameter ($Y$). In the company of feature data, a predictive regression model can be built for future demand as $\hat{Y}=\Psi(\Tilde{X},\hat{\beta}^*)$ where $\hat{\beta}^*=\argmin\limits_{\beta \in \mathbb{R}^{p}} (\Tilde{Y}-\Psi(\Tilde{X},\beta))^T(\Tilde{Y}-\Psi(\Tilde{X},\beta))$ in order to minimize cost function.
However, the training criteria of this predictive regression model will be based on closeness between true and predicted responses via a loss function, the predictive regression model will not capture the effect of holding and backordering costs in the cost function, and future predictions will most probably fail in prescription stage. In Figure \ref{plot:graph1}\subref{plot:graph11}, one predictive regression and two different prescriptive regression models (considering holding and backordering costs) are presented based on two different scenario. When backordering cost is greater than holding cost, using prescriptive regression 1 gives a lower cost on the average since it keeps predictions higher in order to avoid shortage cost as seen in Figure \ref{plot:graph1}\subref{plot:graph12}. Similarly, when holding cost is greater than backordering cost, using prescriptive regression 2 gives a lower cost on the average since it keeps predictions lower in order to avoid holding cost as seen in Figure \ref{plot:graph1}\subref{plot:graph13}. The question is how to obtain such a predictive model directly caring the characteristics of prescriptive model.\\
In this paper, our purpose is to build a framework for predictive regression model caring the characteristics of prescriptive model and providing best decisions. We will call our framework as "Integrated Predictive and Prescriptive Optimization" (IPPO). While IPPO is seeking the most accurate predictive regression model, the desired predictive regression model will tackle uncertainty by providing best actions in prescription stage.
\section{Related Work}
There have been recent developments in supervised machine learning algorithms (classification and regression tasks), but no matter what kind of learning algorithm is employed, the common and final task is generally to train these algorithms in order to find the closest predictions to true values. 
Latest works in the literature show that scientists are realizing that training machine learning models solely based on a prediction criterion like  mean squared error, mean absolute error, or log-likelihood loss is not enough when these noisy predictions will be parameters of prescription tasks. Bengio \cite{Bengio1997UsingAF} is one of the first works considering this issue and building an integrated framework for both prediction and prescription tasks, he emphasized the importance of evaluation criteria of predictive tasks. His neural network model was not designed to minimize prediction error, but instead to  maximize a financial task where those noisy predictions are used as input. 
Another integrated framework developed by Kao et al.\cite{NIPS2009_3686} trains parameters of regression model based on an unconstrained optimization problem with quadratic cost function. It is a hybrid algorithm between ordinary least square and empirical optimization. Tulabandhula and Rudin \cite{Tulabandhula2011MachineLW} minimizes a weighted combination of prediction error and operational cost, but operational cost must be assessed based on true parameters. Bertsimas and Kallus \cite{doi:10.1287/mnsc.2018.3253} add a new dimension to this field by introducing the conditional stochastic optimization term. This simple but efficient idea leverages non-parametric machine learning methods in order to assign weights into train data points for a given test data point, then calculates optimal decisions over stochastic optimization based on calculated weights. However, this methodology is using machine learning tools outside of prescription problem. 
A different approach developed by Ban and Rudin \cite{Ban2019TheBD} considers the decision variables as a function of auxiliary data, but this method may fail if the connection between optimal decisions and parameters in a constrained optimization problem, or may end up with indefeasible decisions. Oroojlooyjadid et al.\cite{Oroojlooyjadid2016ApplyingDL} and Zhang and Gao \cite{Zhang2017AssessingTP} built an extension of \cite{Ban2019TheBD} since both papers approach to the solution from the same perspective, but they solve the problem via neural network to capture the non-linearity between auxiliary data and decisions. 
One more neural network based integrated task is proposed by Donti et al. \cite{NIPS2017_7132}, and primarily they focus on quadratic stochastic optimization problems since they are tuning neural network parameters by differentiating the optimization solution to a stochastic programming problem.
One of the latest works in integrating predictive and prescriptive tasks is developed by Elmachtoub and Grigas  \cite{Elmachtoub2017SmartT} which aims to find parameter of linear regression model inside of decision problem via a new loss function. This method minimizes the difference between objective value provided by true parameters and objective value where decisions provided by predicted parameters are assessed.

\section{Integrated Predictive and Prescriptive Optimization}
\begin{figure}[ht]
\centering
\includegraphics[width=8.45cm, height=3.79cm]{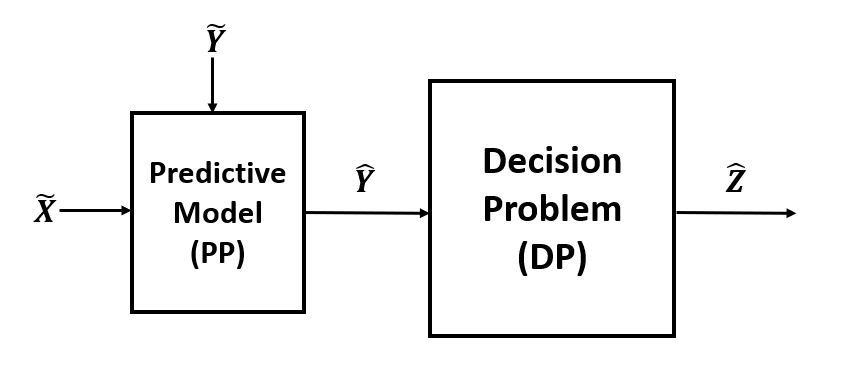}
\caption{Independent Framework}
\label{plot:graph2_1}
\end{figure}

\begin{figure}[ht]
\centering
\includegraphics[width=13.63cm, height=6.11cm]{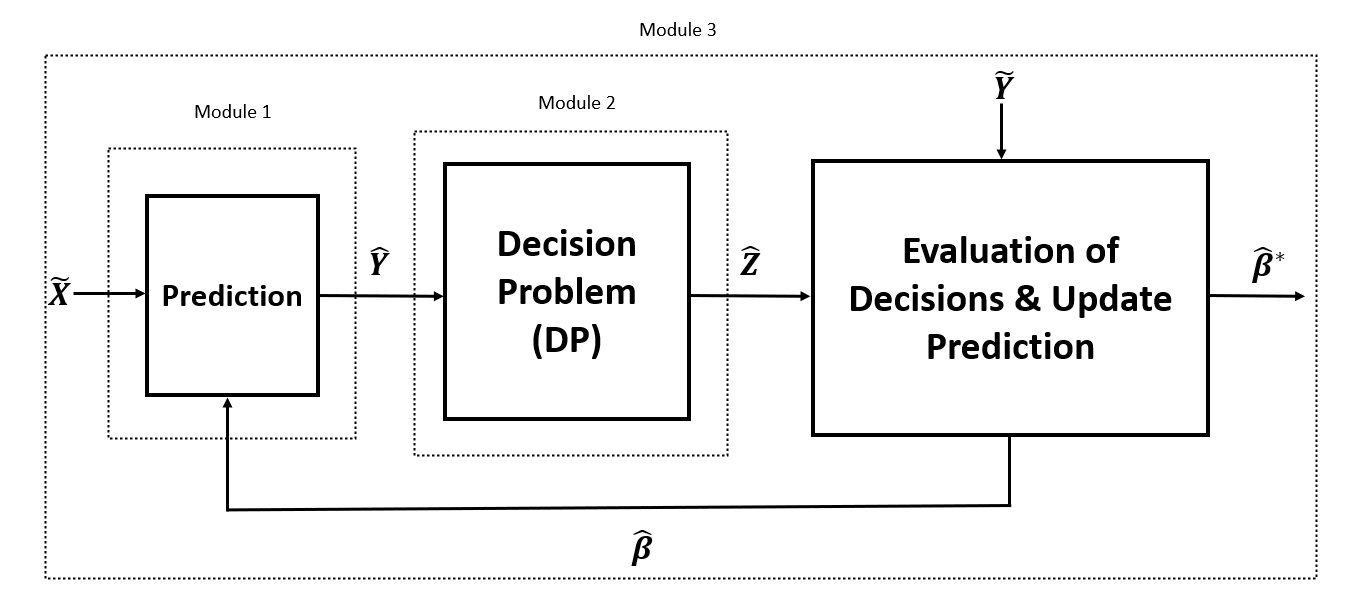}
\caption{Integrated Framework}
\label{plot:graph2_2}
\end{figure}

Given a decision problem (DP) with parameter uncertainty, our goal is to estimate the predictive relationship between responses (uncertain parameters of the decision problem) and a set of input features such that the prescriptive modeling of the decision problem using the predicted responses results in the best decisions. We consider decision problems that can be formulated as an optimization model. Further, the statistical relationship between uncertain parameters (Y) and input features (X) can be approximated through a parametric regression model, i.e., $Y=\Psi(X,\beta)+\epsilon$, where $\beta$ is a vector of $k$ parameters and $\epsilon$ is an error term and $\Psi(\cdot)$ is some function describing the relationship between Y and X. Without loss of generality, given the estimates $\hat{Y}=(\hat{y}_o,\hat{y}_c)$ of uncertain parameters $Y=(y_o,y_c)$, we define the deterministic decision problem (DP) as follows:
\begin{subequations}
\begin{flalign}
\min\limits_{{z \in Z}} \hspace{2 mm} & f(z;\hat{y}_o) \label{eq:eq1-4}\\
\textrm{s.t.} \hspace{2 mm} & g_i(z;\hat{y}_c)\leq 0 \hspace{20 mm} \forall i \in I \hspace{5 mm} c_1\\
& h_j(z;\hat{y}_c)= 0 \hspace{20 mm} \forall j \in J \hspace{3.5 mm} c_2
\end{flalign}
\end{subequations}
In the above formulation, $z$ denote the decision variables, and $\hat{Y}=(\hat{y}_o,\hat{y}_c)$ denote the estimates of the uncertain parameters in the objective and constraints set, respectively. Let ($\hat{z}$) denote the optimal solution of DP given $\hat{Y}=(\hat{y}_o,\hat{y}_c)$, i.e. $\hat{z}=\argmin\{f(z;\hat{y}_o)|z \in Z$ satisfying $c_1$ and $c_2$\}. The uncertain parameters in DP are estimated through a parametric regression model. Let’s assume given the historical data $D=\{(\Tilde{X}^n,\Tilde{Y}^n)\}^N_{n=1}=(\Tilde{X},\Tilde{Y})$ for the response Y and explanatory features X, the prediction problem (PP) using parametric regression is expressed as:
\begin{flalign}
\hat{\beta}^* & =\argmin\limits_{\beta \in \mathbb{R}^{p}} (\Tilde{Y}-\Psi(\Tilde{X},\beta))^T(\Tilde{Y}-\Psi(\Tilde{X},\beta)) \label{eq:eq1-5}
\end{flalign}
In classical approach, the predictive (PP) and prescriptive (DP) tasks which are often treated independently and often in a sequence, i.e., first predict $\hat{Y}=\Psi(\Tilde{X},\hat{\beta}^*)$ (using PP) and then prescribe (using DP). This process is illustrated in Figure \ref{plot:graph2_1}.\\
\newpage
We herein develop an integrated framework joining the predictive (PP) and prescriptive (DP). Three modules of the integration framework are as follows:
\begin{enumerate}
\item Given a set of independent features, a predictive regression model generating responses which are input to the optimization model as part of the input parameter set (Module 1),
\item An optimization model prescribing decisions based on input parameters (Module 2),
\item Another optimization model evaluating the quality of prescribed decisions with respect to ground truth in the response space and updating the parameters of the predictive model (Module 3).
\end{enumerate}
Figure \ref{plot:graph2_2} illustrates these three modules. The sequential predictive and prescriptive tasks (modules 1 and 2) are concurrently optimized through the module 3. While module 1 is a prediction model, modules 2 and 3 are decision optimization problems with their respective decisions influencing one another. The embedding structure of modules 1 and 2 within module 3 is similar to those of bilevel optimization problems.\\
Hence, we model the integration framework as a nested optimization model. In the next section we model the integrated prediction and prescription problem as a bilevel optimization model.

\subsection{Bilevel Models and Solving Techniques}
Bilevel problems are nested optimization problems where an upper-level optimization problem is constrained by a lower-level optimization problem \cite{Bard1998PracticalBO}. A common application of the bilevel problems is a static leader-follower game in economics \cite{Stackelberg}, where the upper level decision maker (leader) has complete knowledge of the lower level problem (follower). Decision variables of the upper level serve as parameters of the lower level. 

We model the integrated framework for the prediction and prescription tasks as a bilevel optimization problem. The upper-level problem jointly determines the parameters of the regression model ($\beta$) and prescription decisions. At lower level, we make decisions with the help of predicted parameters, and we evaluate these actions with respect to true parameters by fixing those "here-and-now" decisions ($Z_f^l$) at upper level problem in constraint \ref{eq:eq1-3}, so that we integrate all proposed steps in one framework formulated as in (\ref{eq:eq1-1})-(\ref{eq:eq1-2}).
\begin{subequations}
\begin{flalign}
\min\limits_{{Z_f^u,Z_s^u,\beta}} \hspace{2 mm} & F(Z_f^u,Z_s^u) \label{eq:eq1-1}\\
\textrm{s.t.} \hspace{2 mm} & G_i(Z_f^u,Z_s^u;Y)\leq 0 \hspace{49 mm} \forall i \in I\\
& H_j(Z_f^u,Z_s^u;Y)=0 \hspace{49 mm} \forall j \in J\\
& Z_f^u=Z_f^l \label{eq:eq1-3}\\
&\min\limits_{Z_f^l,Z_s^l}  \hspace{3 mm} F(Z_f^l,Z_s^l)\\
&\hspace{2.5 mm} \textrm{s.t.} \hspace{3.5 mm} G_i(Z_f^l,Z_s^l;\hat{Y}=\psi(\Tilde{X},\beta))\leq 0 \hspace{19 mm} \forall i \in I\\
&\hspace{12 mm} H_j(Z_f^l,Z_s^l;\hat{Y}=\psi(\Tilde{X},\beta))= 0 \hspace{18 mm} \forall j \in J \label{eq:eq1-2}
\end{flalign}
\end{subequations}

The most popular solution technique for bilevel optimization problems is to transform bilevel problem into single level problem by replacing objective function of lower level problem with Karush–Kuhn–Tucker (KKT) conditions \cite{Bard1998PracticalBO}. The KKT conditions appear as dual and complementarity constraints, that is why KKT conditions require convexity, so this approach is limited to convex lower level problems. Complementarity constraints (\ref{eq:eq2-2} and \ref{eq:eq2-3})  convert the problem nonlinear models; thus, these constraints are replaced with logic constraints by defining new binary variables and sufficiently enough an M parameter. With this final touch, bilevel model turns into a mixed-integer problem, and it can be solved by traditional solvers as in form (\ref{eq:eq8})-(\ref{eq:eq9}).\\
\vspace{-10mm}
\begin{subequations}
\begin{flalign}
\min\limits_{\substack{Z_f^u,Z_s^u,Z_f^l,Z_s^l; \\ \beta,\pi_m,\mu_n}} \hspace{2 mm} & F(Z_f^u,Z_s^u) \label{eq:eq8}\\
\textrm{s.t.} \hspace{8 mm} & G_i(Z_f^u,Z_s^u;Y)\leq 0 \hspace{22 mm} \forall i \in I\\
& H_j(Z_f^u,Z_s^u;Y)= 0 \hspace{21.5 mm} \forall j \in J\\
& Z_f^u=Z_f^l \hspace{39 mm} \label{eq:eq2-1}\\
& G_i(Z_f^l,Z_s^l;\hat{Y})\leq 0 \hspace{23 mm} \forall i \in I\\
& H_j(Z_f^l,Z_s^l;\hat{Y})= 0 \hspace{22 mm} \forall j \in J\\
& \nabla_{Z_f^l,Z_s^l} \hspace{1 mm} L(Z_f^l,Z_s^l,\pi_i,\mu_j)=0\\
& G_i(Z_f^l,Z_s^l;\hat{Y})\pi_i =0 \hspace{20 mm} \forall i \in I \label{eq:eq2-2}\\
& H_j(Z_f^l,Z_s^l;\hat{Y})\mu_j =0 \hspace{19 mm} \forall j \in J \label{eq:eq2-3}\\
& \pi_i \geq 0,\quad \mu_j \geq 0 \hspace{26.2 mm} \forall i \in I,\quad \forall j \in J \label{eq:eq9}\\
where \hspace{5 mm}& \nonumber \\
& L(Z_f^l,Z_s^l,\pi_i,\mu_j)=F(Z_f^l,Z_s^l)+\sum\limits_{i=1}^{I} \pi_i G_i(Z_f^l,Z_s^l;\hat{Y})+\sum\limits_{j=1}^{J} \mu_j H_j(Z_f^l,Z_s^l;\hat{Y}) \nonumber\\
& \hat{Y}=\psi(\Tilde{X},\beta) \nonumber
\end{flalign}
\end{subequations}
\newpage
\subsection{Controlling Generalization Error in IPPO}
\label{sec:ControllingGeneralizationErrorinIPPO}
Our model is developed based on finding best decision in train data set. In order to ensure quality of prediction model in external data set, we propose three different ways to control generalization error within this framework by regularizing predictive model parameters. The first method is to rewrite objective function as weighted average of predictive error and prescriptive error as shown in formulation (\ref{eq:eq3-1})-(\ref{eq:eq3-2}) where $0 \leq \lambda_1 \leq 1$. When $\lambda_1=1$, we solve pure bilevel optimization model without generalization error, and when $\lambda_1=0$, we ignore the prescription part, and optimize directly predictive algorithm solely, and that leads us to point estimate-based prescriptions.
\begin{subequations}
\begin{flalign}
\min\limits_{{Z_f^u,Z_s^u,\beta}} \hspace{2 mm} & \lambda_1F(Z_f^u,Z_s^u)+(1-\lambda_1) L(Y,\hat{Y}=\psi(\Tilde{X},\beta)) \label{eq:eq3-1}\\
\textrm{s.t.} \hspace{2 mm} & G_i(Z_f^u,Z_s^u;Y)\leq 0 \hspace{49 mm} \forall i \in I\\
& H_j(Z_f^u,Z_s^u;Y)=0 \hspace{49 mm} \forall j \in J\\
& Z_f^u=Z_f^l\\
&\min\limits_{Z_f^l,Z_s^l}  \hspace{3 mm} F(Z_f^l,Z_s^l)\\
&\hspace{2.5 mm} \textrm{s.t.} \hspace{3.5 mm} G_i(Z_f^l,Z_s^l;\hat{Y}=\psi(\Tilde{X},\beta))\leq 0 \hspace{19 mm} \forall i \in I\\
&\hspace{12 mm} H_j(Z_f^l,Z_s^l;\hat{Y}=\psi(\Tilde{X},\beta))= 0 \hspace{18 mm} \forall j \in J \label{eq:eq3-2}
\end{flalign}
\end{subequations}
The second method uses predictive error term again, but in a way that it can be restricted by a constraint. However this restriction cannot be less than the loss value which is provided by predictive model solely because constraining loss less than the optimal ($L^*$) makes optimization model indefeasible, this method is shown in formulation (\ref{eq:eq4-1})-(\ref{eq:eq4-2}) with a restriction parameter $\lambda_2 \geq 1$.\\
\begin{subequations}
\begin{flalign}
\min\limits_{{Z_f^u,Z_s^u,\beta}} \hspace{2 mm} & F(Z_f^u,Z_s^u) \label{eq:eq4-1}\\
\textrm{s.t.} \hspace{2 mm} & G_i(Z_f^u,Z_s^u;Y)\leq 0 \hspace{49 mm} \forall i \in I\\
& H_j(Z_f^u,Z_s^u;Y)=0 \hspace{49 mm} \forall j \in J\\
& L(Y,\hat{Y}=\psi(\Tilde{X},\beta)) \leq \lambda_2L^*\\
& Z_f^u=Z_f^l\\
&\min\limits_{Z_f^l,Z_s^l}  \hspace{3 mm} F(Z_f^l,Z_s^l)\\
&\hspace{2.5 mm} \textrm{s.t.} \hspace{3.5 mm} G_i(Z_f^l,Z_s^l;\hat{Y}=\psi(\Tilde{X},\beta))\leq 0 \hspace{19 mm} \forall i \in I\\
&\hspace{12 mm} H_j(Z_f^l,Z_s^l;\hat{Y}=\psi(\Tilde{X},\beta))= 0 \hspace{18 mm} \forall j \in J \label{eq:eq4-2}
\end{flalign}
\end{subequations}
The last method is to shrink predictive model parameters by penalizing with a penalty coefficient $\lambda_3 \geq 0$ as \cite{Tibshirani94regressionshrinkage} introduced, this model is formulated in (\ref{eq:eq5-1})-(\ref{eq:eq5-2}).\\
\begin{subequations}
\begin{flalign}
\min\limits_{{Z_f^u,Z_s^u,\beta}} \hspace{2 mm} & F(Z_f^u,Z_s^u)+\lambda_3 \beta^T \beta \label{eq:eq5-1}\\
\textrm{s.t.} \hspace{2 mm} & G_i(Z_f^u,Z_s^u;Y)\leq 0 \hspace{49 mm} \forall i \in I\\
& H_j(Z_f^u,Z_s^u;Y)=0 \hspace{49 mm} \forall j \in J\\
& Z_f^u=Z_f^l\\
&\min\limits_{Z_f^l,Z_s^l}  \hspace{3 mm} F(Z_f^l,Z_s^l)\\
&\hspace{2.5 mm} \textrm{s.t.} \hspace{3.5 mm} G_i(Z_f^l,Z_s^l;\hat{Y}=\psi(\Tilde{X},\beta))\leq 0 \hspace{19 mm} \forall i \in I\\
&\hspace{12 mm} H_j(Z_f^l,Z_s^l;\hat{Y}=\psi(\Tilde{X},\beta))= 0 \hspace{18 mm} \forall j \in J \label{eq:eq5-2}
\end{flalign}
\end{subequations}

\subsection{Proposed Decomposition Method for IPPO}
Bilevel optimization problems are NP-hard, and it is not easy to solve. However, our proposed predictive and prescriptive integrated methodology has a unique feature. All the defined variables belong to its own scenario except the regression parameters. Regression parameters are common for all scenarios. After converting bilevel to a single level problem by applying KKT conditions, our model becomes a two-stage mixed-integer program whose first stage variables are regression parameters. Here, we create copies of regression parameters across all scenarios and make the problem fully scenario-based decomposable, but we need to include a non-anticipativity or implementability constraint to ensure all regression parameters are equal each other for all scenarios. Progressive hedging algorithm (PHA) proposed by \cite{RockWets91} can be used as decomposition techniques for our two stage mixed-integer problem. In our framework, we will provide best candidate solution as initial regression parameters for depicted Figure \ref{plot:graph2_2}, and PHA solves all scenario problems independently, then we wıll update regression parameters ıteratıvely. These steps repeat until a convergence is satisfied.
\begin{subequations}
\begin{flalign}
\min\limits_{z^f \in Z^F,z^s \in Z^S} \hspace{2 mm} & c^fz^f+\sum\limits_{i=1}^{I}c_i^sz_i^s \label{eq:eq10}\\
\textrm{s.t.} \hspace{9 mm} & Az^f \geq b \label{eq:eq11}\\
& T_iz^f+W_iz_i^s \geq r_i \hspace{21.6 mm} \forall i \in I \label{eq:eq12}
\end{flalign}
\end{subequations}
For better understanding, let’s consider the formulation in (\ref{eq:eq10})-(\ref{eq:eq12}), $z^f$ indicates first stage variable, and $z^s_i$ indicates second stage variable for $i^{th}$ scenario.
\begin{subequations}
\begin{flalign}
\min\limits_{\substack{z^f \in Z^F,z^s \in Z^S;\\z^d \in Z^D}} \hspace{2 mm} & \sum\limits_{i=1}^{I}(c^fz^d_i+c_i^sz_i^s) \label{eq:eq13}\\
\textrm{s.t.} \hspace{9 mm} & Az_i^d \geq b \hspace{37 mm} \forall i \in I \label{eq:eq14}\\
& T_iz^d_i+W_iz_i^s \geq r_i \hspace{21.6 mm} \forall i \in I \label{eq:eq15}\\
& z_i^d-z^f=0 \hspace{31 mm} \forall i \in I \label{eq:eq16}
\end{flalign}
\end{subequations}
In the formulation  (\ref{eq:eq13})-(\ref{eq:eq16}), first stage variable is duplicated, and $z^d_i$ variables are created for each scenario, but they are linked via non-anticipativity constraint \ref{eq:eq16}. By relaxing constraint \ref{eq:eq16}, all scenarios can be easily solved in a parallel way. 

PHA iterates and converges to a common solution taking into account all the scenarios belonging to the original problem. We show the details and steps for basic PHA in Algorithm \ref{algo:algo1}. Let $\rho>0$ be penalty factor, $\delta$ be stopping criteria, and $W$ be dual prices for non-anticipativity constraint \ref{eq:eq16}.\\
\begin{algorithm}
\textbf{Initialization} \linebreak
\linebreak
$k=0$ \linebreak
\linebreak
$z^{d,k}_i=\argmin\limits_{z^d \in Z^D,z^s \in Z^S}(c^fz^d_i+c_i^sz_i^s)$ \quad s.t. (\ref{eq:eq14})-(\ref{eq:eq15})\hspace{5 mm} $\forall i \in I$ \linebreak
\linebreak
$\overline{z}^k=\frac{\sum\limits_{i=1}^{I}z^{d,k}_i}{I}$ \linebreak
\linebreak
$w_i^k=\rho(z^{d,k}_i-\overline{z}^k)$  \hspace{5 mm} $\forall i \in I$ \linebreak
\\
\textbf{Iteration Update}\linebreak
\linebreak
$k=k+1$\linebreak
\\
\textbf{Decomposition}\linebreak
\linebreak
$z^{d,k}_i=\argmin\limits_{z^d \in Z^D,z^s \in Z^S}(c^fz^d_i+c_i^sz_i^s+w_i^{k-1}z^d_i+\frac{\rho}{2}(z^d_i-\overline{z}^{k-1})^2$ \quad s.t. (\ref{eq:eq14})-(\ref{eq:eq15}) \hspace{5 mm} $\forall i \in I$ \linebreak
\linebreak
$\overline{z}^k=\frac{\sum\limits_{i=1}^{I}z^{d,k}_i}{I}$ \linebreak
\linebreak
$w_i^k=w_i^{k-1}+\rho(z^{d,k}_i-\overline{z}^k)$  \hspace{5 mm} $\forall i \in I$ \linebreak
\\
\textbf{Convergence Check}\linebreak
\linebreak
If all scenario solutions $z^{d,k}_i$ are equal with at most $\delta$ deviation, stop. Else, go to step 2.\linebreak
\\
\caption{The Progressive Hedging Algorithm}
\label{algo:algo1}
\end{algorithm}

\section{Experimental Setup}
In this part, we discuss why and how we select predictive and prescriptive model, then we will introduce parameters and variables of these two tasks. Next, we explain data creation process step by step, and we will show formulation of integrated predictive and prescriptive task. 

\subsection{Prescriptive and Predictive Model Selection}
We validate the performance of integrated predictive and prescriptive methodology and compare it with various well-known and recently developed methods. We perform numerical experiments on two different prescriptive models. The first one is well-known newsvendor problem used by \cite{Ban2019TheBD}, but we extend it from single product to multi-product newsvendor problem ($d_l=12$ products), and we use different costs for each scenario (production , holding, and backordering) instead of fix costs in order to increase the complexity of problem. 
Extensive form of classical newsvendor problem with multi-product is expressed as formulated in (\ref{eq:eq2})-(\ref{eq:eq17}).
\begin{subequations}
\begin{flalign}
\min\limits_{Q,U,O} \hspace{5 mm} & \frac{1}{J}\sum\limits_{j=1}^{J}\left[ \frac{1}{N}\sum\limits_{n=1}^{N}(c_{n,j}Q_j+b_{n,j}U_{n,j}+h_{n,j}O_{n,j})\right] \label{eq:eq2}\\
\textrm{s.t.} \hspace{5 mm} & U_{n,j} \geq Y_{n,j}-Q_j \hspace{30 mm} \forall j \in J, \forall n \in N\\
&O_{n,j} \geq Q_j-Y_{n,j} \hspace{30 mm} \forall j \in J, \forall n \in N\\
&Q_{j}, U_{n,j},O_{n,j}\geq 0 \hspace{29 mm} \forall j \in J, \forall n \in N \label{eq:eq17}
\end{flalign}
\end{subequations}
\begin{tabular}{ll}
\multicolumn{2}{l}{\textit{\textbf{Decision Variables for Newsvendor Problem}}} \\
$Q_j$&Amount of regular order done in advance for product $j$\\
$U_{n,j}$&Amount of shortage for product $j$ in scenario $n$\\
$O_{n,j}$&Amount of surplus for product $j$ in scenario $n$\\
\multicolumn{2}{l}{} \\
\multicolumn{2}{l}{\textit{\textbf{Parameters  for Newsvendor Problem}}} \\
$c_{n,j}$&Cost of order for product $j$ in scenario $n$\\
$b_{n,j}$&Cost of backordering for product $j$ in scenario $n$\\
$h_{n,j}$&Cost of hold for product $j$ in scenario $n$\\
$Y_{n,j}$&Amount of demand for product $j$ in scenario $n$\\
\end{tabular}\\

The second prescriptive model is two-stage shipment planning problem leveraged by \cite{doi:10.1287/mnsc.2018.3253} where there is a network between $d_w=4$ warehouses and $d_l=12$ locations. The goal is to produce and hold a product at a cost in warehouses to satisfy the future demand of locations. Then the product is shipped, when needed, from warehouses to locations with transportation cost. In case current total supply in warehouses does not satisfy the demand of locations, the last-minute production takes place with a higher cost. The extensive form of two-stage shipment problem is formulated as follow in (\ref{eq:eq18})-(\ref{eq:eq19}).

\begin{subequations}
\begin{flalign}
\min\limits_{Z,T,S} \hspace{5 mm} & \sum\limits_{i=1}^{I}P_1Z_i+\frac{1}{N}\sum\limits_{n=1}^{N}\left[\sum\limits_{i=1}^{I}P_2T_{n,i}+\sum\limits_{i=1}^{I}\sum\limits_{j=1}^{J}C_{n,i,j}S_{n,i,j}\right] \label{eq:eq18}\\
\textrm{s.t.} \hspace{5 mm} & \sum\limits_{i=1}^{I}S_{n,i,j}\geq Y_{n,j} \hspace{31 mm} \forall j \in J,\forall n \in N\\
&\sum\limits_{j=1}^{J}S_{n,i,j}\leq Z_{i}+T_{n,i} \hspace{22 mm} \forall i \in I,\forall n \in N\\
&Z_{i},\quad T_{n,i},\quad S_{n,i,j}\geq 0 \hspace{21 mm} \forall i \in I,\forall j \in J,\forall n \in N \label{eq:eq19}
\end{flalign}
\end{subequations}
\begin{tabular}{ll}
\multicolumn{2}{l}{\textit{\textbf{Decision Variables}}} \\
$Z_{i}$&Amount of production done in advance at warehouse $i$\\
$T_{n,i}$&Amount of production done last minute at warehouse $i$ in scenario $n$\\
$S_{n,i,j}$&Amount of shipment from warehouse $i$ to location $j$ in scenario $n$\\
\multicolumn{2}{l}{} \\
\multicolumn{2}{l}{\textit{\textbf{Parameters}}} \\
$P_1$&Cost of production done in advance at warehouse\\
$P_2$&Cost of production done last minute at warehouses\\
$C_{n,i,j}$&Cost of shipment from warehouse $i$ to location $j$ in scenario $n$\\
$Y_{n,j}$&Amount of demand at location $j$ in scenario $n$\\
\end{tabular}\\

As for predictive model, since we embed the predictive model inside of prescriptive model, we choose linear regression model as in \ref{eq:eq20} to maintain the linearity of prescriptive model. However, other predictive methodologies still can be applied to capture nonlinearity outside of this integrated framework as preprocess, and dimensionality can be reduced between feature variables and responses especially in high dimensional data as built in deep learning, then these converted feature variables can be embedded inside of prescriptive model via linear regression again as described above.
\begin{flalign}
\hat{Y}_{n,j} & =\beta_{j,0}X_{n,0}+\sum\limits_{f=1}^{F}\beta_{j,f}X_{n,f} \quad \forall j \in J \quad \forall n \in N \quad where \quad X_{n,0}=1 \label{eq:eq20}
\end{flalign}

\subsection{Data Generation}
\label{subsec:DataGeneration}
In both experiments, we randomly generate feature variables of predictive model based on a $d_x=3$ dimensional multivariate normal distribution with size of $n=2000$ observations, $X \in \mathbb{R}^{n \times d_x}$, i.e., $X\sim N(\mu, \Sigma)$, where $\mu=[0,0,0]$ and $\Sigma=[[1,0.5,-0.5],[0.5,1,-0.5],[-0.5,0.5,1]]$. Then, we choose the true parameters of our predictive model, linear regression in our case, as $\beta \in \mathbb{R}^{d_l \times d_x}$ matrix for slopes and intercepts. Next, we calculated response according to the model $Y=\beta_0+\beta X^T+\varepsilon$, where $\varepsilon$ is independently generated noise term and follows normal distribution, i.e., $\varepsilon \sim N(0, \sigma)$. Here the standard deviation of added noise controls the correlation between feature values and responses (Responses represent demand is in both prescriptive problems). To see the behavior of our method and other methods, we have employed 10 different noise standard deviations, thus we create 10 different feature and response pairs with different correlations. we measure these correlations based on R-Square value of a linear regression model. As for shipment cost, we randomly simulate its matrix as from warehouse $i$ to location $j$ based on uniform distribution $C_{n,i,j} \sim U(0,30)$ for each scenario. In newsvendor problem. we create order, backordering, and holding costs again based on uniform distribution $c_{n,j} \sim U(0,300)$, $b_{n,j} \sim U(0,3000)$, and $h_{n,j} \sim U(0,150)$ for each scenario and product, respectively. Out of created $n=2000$ observations, we randomly choose train, validation, and test sets from $n=2000$ observations with size of $70, 15, 15$, respectively. This splitting process is repeated by $30$ times, and all results are reported based on the average cost of these $30$ replications in both problems.
\subsection{IPPO Formulations}
We modify newsvendor and two-stage shipment problems here for integration process. First, we introduce $\beta$ variable to make predictions for demand via linear regression. Then we also introduce counterpart decision variables of original variables in newsvendor and shipment models because these counterpart decision variables will be made based on predicted demands. In our lower level, we make decisions based on output of predictive algorithm, linear regression, and submit these decisions to the upper level, so that we can evaluate the quality of these decisions based on true parameters in \ref{eq:eq24} and \ref{eq:eq33}.\\
\begin{subequations}
\begin{flalign}
\min\limits_{Q^U,U^U,O^U,\beta} & \hspace{3 mm} \frac{1}{J}\sum\limits_{j=1}^{J}\left[\frac{1}{N}\sum\limits_{n=1}^{N}(c_{n,j}Q_{n,j}^{U}+b_{n,j}U_{n,j}^U+h_{n,j}O_{n,j}^U)\right] \label{eq:eq21}\\
\textrm{s.t.} \hspace{5 mm} & U_{n,j}^U \geq Y_{n,j}-Q_{n,j}^U \hspace{47.1 mm} \forall j \in J, \forall n \in N \label{eq:eq22}\\
& O_{n,j}^U \geq Q_{n,j}^U-Y_{n,j} \hspace{47 mm} \forall j \in J, \forall n \in N \label{eq:eq23}\\
& Q^U_{n,j}= Q^L_{n,j}   \hspace{58.3 mm} \forall j \in J,\forall n \in N \label{eq:eq24}\\
& Q_{n,j}^U, U_{n,j}^U,O_{n,j}^U\geq 0, \quad \beta \quad free \hspace{25.1 mm} \forall j \in J, \forall n \in N \label{eq:eq25}\\
&\min\limits_{Q^L,U^L,O^L}  \frac{1}{J}\sum\limits_{j=1}^{J}\left[ \frac{1}{N}\sum\limits_{n=1}^{N}(c_{n,j}Q_{n,j}^{L}+b_{n,j}U_{n,j}^L+h_{n,j}O_{n,j}^L)\right] \label{eq:eq26}\\
&\hspace{6 mm} \textrm{s.t.} \hspace{4 mm} U_{n,j}^L \geq \hat{Y}_{n,j}-Q_{n,j}^L \hspace{32 mm} \forall j \in J, \forall n \in N \label{eq:eq27}\\
&\hspace{16 mm} O_{n,j}^L \geq Q_{n,j}^L-\hat{Y}_{n,j} \hspace{31.5 mm} \forall j \in J, \forall n \in N \label{eq:eq28}\\
&\hspace{16 mm} Q_{n,j}^L, U_{n,j}^L,O_{n,j}^L\geq 0 \hspace{31 mm} \forall j \in J,\forall n \in N \label{eq:eq29}
\end{flalign}
\end{subequations}
We modify newsvendor and two-stage shipment problems here for integration process. First, we introduce $\beta$ variable to make predictions for demand via linear regression. Then we also introduce counterpart decision variables of original variables in newsvendor and shipment models because these counterpart decision variables will be made based on predicted demands. In our lower level, we make decisions based on output of predictive algorithm, linear regression, and submit these decisions to the upper level, so that we can evaluate the quality of these decisions based on true parameters. Constraints \ref{eq:eq24} and \ref{eq:eq33} undertake this task.\\
Detailed formulation of integrated newsvendor problem and  two stage shipment problem is provided below in (\ref{eq:eq21})-(\ref{eq:eq29}) and (\ref{eq:eq30})-(\ref{eq:eq38}), respectively. If controlling generalization error is needed, one of the recommendations formulated in \autoref{sec:ControllingGeneralizationErrorinIPPO} can be included in these models.\\
In both formulations, there will be a trade-off between lower and upper problems, such that lower problem minimizes its own cost based on $\beta$ variables provided by upper level problem, and upper level problem minimizes its own objective value based on prescriptions provided by lower level problem. To be able to solve bilevel model, we need to add optimality conditions for lower level problem based on KKT conditions. We introduce dual variables for lower level constraints, and write these conditions, but KKT conditions bring nonlinearity because of complementary slackness, so new binary variables can be defined, and SOS constraints and big M method can be used. After introducing KKT conditions, predictive task integrated two stage shipment problem becomes a mix integer problem with single level, and this formulation is given in (\ref{eq:eq8})-(\ref{eq:eq9}).
\begin{subequations}
\begin{flalign}
\min\limits_{\beta,Z^U,T^U,S^U} & \frac{1}{N}\sum\limits_{n=1}^{N}(\sum\limits_{i=1}^{I}P_1Z^U_{n,i}+\sum\limits_{i=1}^{I}P_2T^U_{n,i}+\sum\limits_{i=1}^{I}\sum\limits_{j=1}^{J}C_{n,i,j}S^U_{n,i,j}) \label{eq:eq30}\\
\textrm{s.t.} \hspace{5 mm} & \sum\limits_{i=1}^{I}S^U_{n,i,j}\geq Y_{n,j} \hspace{60 mm} \forall n \in N,\forall j \in J \label{eq:eq31}\\
& \sum\limits_{j=1}^{J}S^U_{n,i,j}\leq Z^U_{n,i}+T^U_{n,i} \hspace{48.5 mm} \forall n \in N,\forall i \in I  \label{eq:eq32}\\
& Z^U_{n,i}= Z^L_{n,i}   \hspace{69.45 mm} \forall n \in N,\forall i \in I \label{eq:eq33}\\
& Z^U_{n,i}, \quad T^U_{n,i}, \quad S^U_{n,i,j}\geq 0,\quad \beta \quad free  \hspace{26 mm} \forall n \in N,\forall i \in I, \forall j \in J \label{eq:e34}\\
&\min\limits_{Z^L,T^L,S^L}  \frac{1}{N}\sum\limits_{n=1}^{N}(\sum\limits_{i=1}^{I}P_1Z^L_{n,i}+\sum\limits_{i=1}^{I}P_2T^L_{n,i}+\sum\limits_{i=1}^{I}\sum\limits_{j=1}^{J}C_{n,i,j}S^L_{n,i,j}) \label{eq:eq35}\\
&\hspace{6 mm} \textrm{s.t.} \hspace{4 mm} \sum\limits_{i=1}^{I}S^L_{n,i,j}\geq \hat{Y}_{n,j}  \hspace{44 mm} \forall n \in N,\forall j \in J \label{eq:eq36}\\
&\hspace{16 mm} \sum\limits_{j=1}^{J}S^L_{n,i,j}\leq Z^L_{n,i}+T^L_{n,i} \hspace{32.5 mm} \forall n \in N,\forall i \in I \label{eq:eq37}\\
&\hspace{16 mm} Z^L_{n,i},\quad T^L_{n,i},\quad S^L_{n,i,j}\geq 0 \hspace{31 mm} \forall n \in N,\forall i \in I, \forall j \in J \label{eq:eq38}
\end{flalign}
\end{subequations}   

\subsection{Convergence to Stochastic Optimization}
The Equation in \ref{eq:eq20} defines the linear regression where the output is a weighted combination of inputs plus an intercept. Linear regressions are generally trained based on mean squared deviations. In our proposed integrated model, this Equation in \ref{eq:eq20} produces predictions for each scenario, and lower level objective and constraints prescribes decisions based on these predictions as seen in (\ref{eq:eq26})-(\ref{eq:eq29}) and (\ref{eq:eq35})-(\ref{eq:eq38}) for each scenario again. If level of correlation between $X$ and $Y$ goes to zero, slopes of predictive model in  \ref{eq:eq20} or the feature variable contributions goes to zero, and predictions will be all equal each other thanks to intercepts. The same prediction for all scenarios will prescribe the same decisions across all scenarios as seen in forwarded decisions from lower level to upper level in (\ref{eq:eq24}) and (\ref{eq:eq33}). This converts our integrated methodology to a single level problem and becomes scenario formulation of a two-stage stochastic problem as shown in (\ref{eq:eq13})-(\ref{eq:eq16}).

\section{Computational Results}
This section discusses the performance of our integrated methodology and other methods under different circumstances. All results for these experiments are obtained from Gurobi python API \cite{gurobi}. We compared results of various well-known methods like Point-Estimate-Based Optimization, Stochastic Optimization, and recent methods like Conditional Stochastic Optimization (kNN), and The Feature-Based Optimization by \cite{doi:10.1287/mnsc.2018.3253}, \cite{Ban2019TheBD}, respectively. We investigate the behaviors of these methods under different correlations between features $X$ and responses $Y$, and evaluate performance of validation data set to see if generalization error is needed.

\subsection{Newsvendor Problem}
\begin{figure}
\makebox[\linewidth][c]{%
\begin{subfigure}[b]{.6\textwidth}
\centering
\includegraphics[width=.95\textwidth]{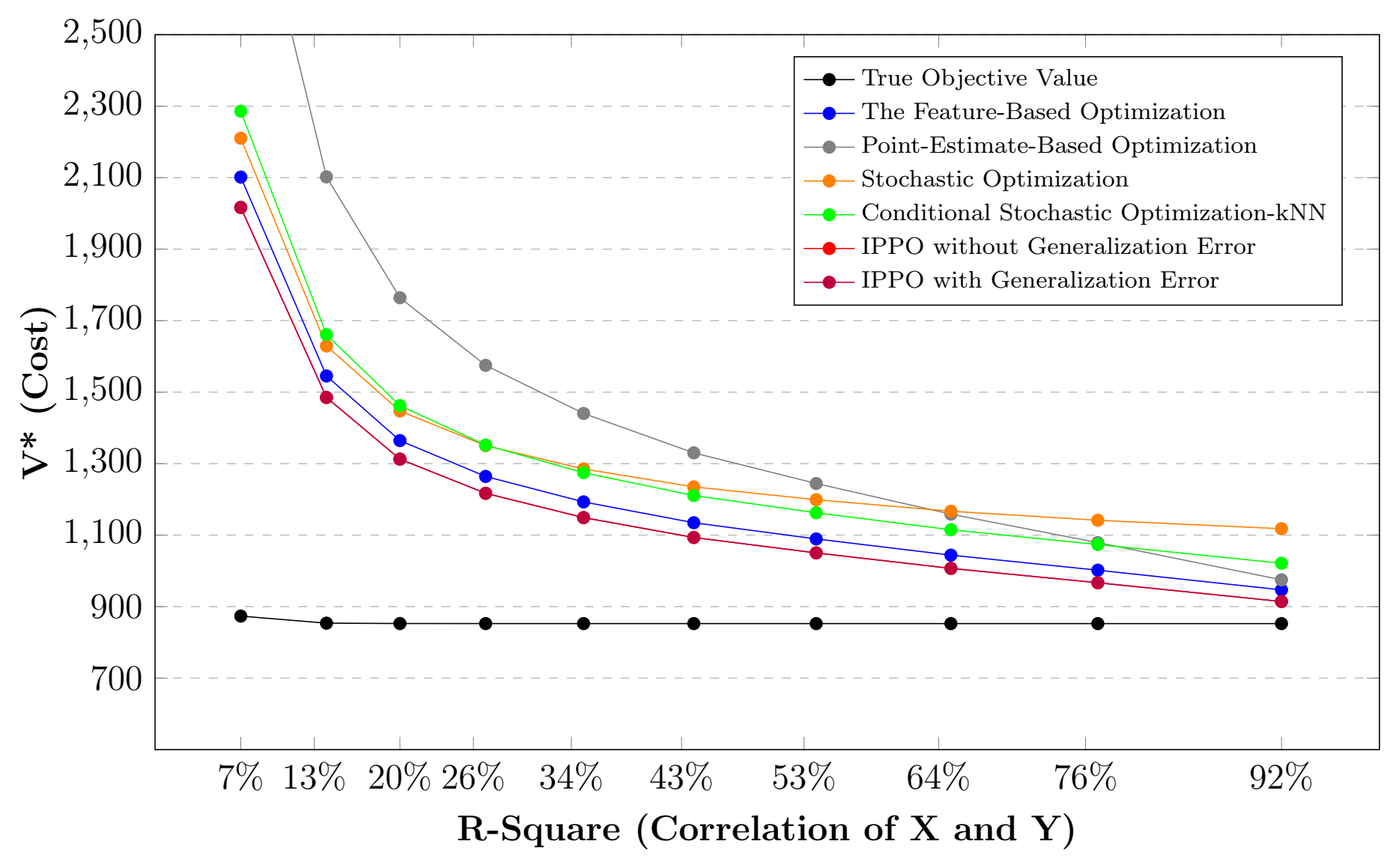}
\caption{Train Data Set}
\end{subfigure} \hspace{-5mm}%
\begin{subfigure}[b]{.6\textwidth}
\centering
\includegraphics[width=.95\textwidth]{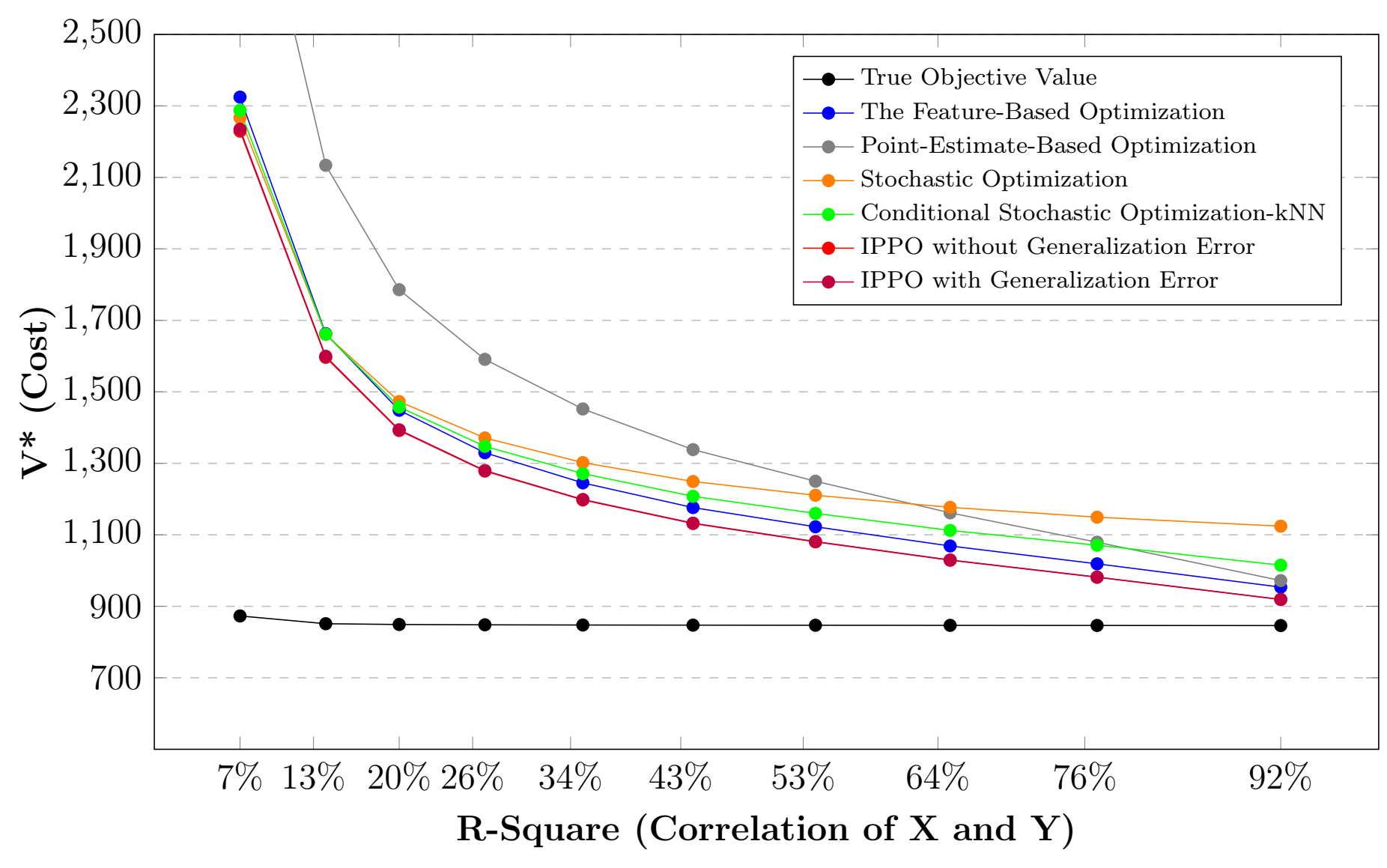}
\caption{Test Data Set}
\end{subfigure}%
}\\
\caption{Comparison of Different Methods for Newsvendor Problem}
\label{plot:graph3}
\end{figure}

First and common feature of all methods, as we see in Figure \ref{plot:graph3}, solution quality improves when we increase the correlation between features (side info) and responses, and this is expected because the more information is provided, the better results are obtained. However, improvement rates are different in each method. Second, kNN from \cite{doi:10.1287/mnsc.2018.3253} gives better result compared to stochastic optimization (k neighbors value is optimized over validation data set). The reason why it gives better solution is because it leverages specific neighbors in train data, instead of minimizing expected cost over all train data, eliminates irrelevant scenarios, search optimal solution by minimizing expected cost around neighbors.

\begin{table}
\resizebox{\textwidth}{!}{\begin{tabular}{|c|c|c|c|c|c|c|c|c|c|c|c|}
\hline
\multirow{2}{*}{\textbf{\begin{tabular}[c]{@{}c@{}}Correlation \\ between X and Y\end{tabular}}} & \multicolumn{4}{c|}{\textbf{True Objective Values}}           & \multicolumn{4}{c|}{\textbf{IPPO}}                            & \multirow{2}{*}{\textbf{Performance}} & \multirow{2}{*}{\textbf{\begin{tabular}[c]{@{}c@{}}Optimal Regularization \\ Parameter\end{tabular}}} & \multirow{2}{*}{\textbf{\begin{tabular}[c]{@{}c@{}}Optimal Neighbors\\ for  kNN\end{tabular}}} \\ \cline{2-9}
                                                                                                 & \textbf{Mean} & \textbf{Max} & \textbf{Min} & \textbf{S.Dev.} & \textbf{Mean} & \textbf{Max} & \textbf{Min} & \textbf{S.Dev.} &                                       &                                                                                                       &                                                                                                \\ \hline
\%7                                                                                                & 873.5         & 924.8        & 815.0        & 27.2            & 2016.3        & 2118.5       & 1897.9       & 58.3            & \%7                                     & 1 & 41                                                                                                    \\ \hline
\%13                                                                                               & 853.7         & 892.9        & 813.6        & 21.0            & 1484.9        & 1549.1       & 1410.3       & 35.0            & \%10                                    & 1& 41                                                                                                     \\ \hline
\%20                                                                                               & 852.7         & 889.5        & 814.5        & 19.6            & 1312.5        & 1369.6       & 1252.7       & 28.3            & \%11                                    & 1   & 31                                                                                                  \\ \hline
\%26                                                                                               & 852.5         & 889.2        & 815.3        & 19.0            & 1216.8        & 1270.1       & 1164.7       & 25.0            & \%13                                    & 1 & 28                                                                                         \\ \hline
\%34                                                                                               & 852.5         & 889.4        & 816.1        & 18.6            & 1148.9        & 1199.4       & 1102.0       & 22.9            & \%15                                    & 1    & 28                                                                                                 \\ \hline
\%43                                                                                               & 852.5         & 889.6        & 816.8        & 18.3            & 1093.3        & 1141.6       & 1050.7       & 21.4            & \%17                                    & 1    & 22                                                                                                 \\ \hline
\%53                                                                                               & 852.5         & 889.8        & 817.3        & 18.2            & 1050.1        & 1096.5       & 1010.9       & 20.4            & \%20                                    & 1    & 16                                                                                                 \\ \hline
\%64                                                                                               & 852.5         & 890.0        & 817.8        & 18.1            & 1006.9        & 1051.5       & 971.0        & 19.5            & \%24                                    & 1    & 16                                                                                                 \\ \hline
\%76                                                                                               & 852.5         & 890.2        & 818.3        & 18.0            & 966.7         & 1009.7       & 933.4        & 18.8            & \%31                                    & 1       & 10                                                                                              \\ \hline
\%92                                                                                               & 852.5         & 890.4        & 818.9        & 17.9            & 914.2         & 955.0        & 881.2        & 18.2            & \%53                                    & 1      & 5                                                                                               \\ \hline
\end{tabular}}
\caption{Newsvendor Problem Train Data Set Statistical Values} 
\label{tab:table1}
\end{table} 
\begin{table}
\resizebox{\textwidth}{!}{\begin{tabular}{|c|c|c|c|c|c|c|c|c|c|c|c|}
\hline
\multirow{2}{*}{\textbf{\begin{tabular}[c]{@{}c@{}}Correlation \\ between X and Y\end{tabular}}} & \multicolumn{4}{c|}{\textbf{True Objective Values}}           & \multicolumn{4}{c|}{\textbf{IPPO}}                            & \multirow{2}{*}{\textbf{Performance}} & \multirow{2}{*}{\textbf{\begin{tabular}[c]{@{}c@{}}Optimal Regularization \\ Parameter\end{tabular}}} & \multirow{2}{*}{\textbf{\begin{tabular}[c]{@{}c@{}}Optimal Neighbors\\ for  kNN\end{tabular}}} \\ \cline{2-9}
                                                                                                 & \textbf{Mean} & \textbf{Max} & \textbf{Min} & \textbf{S.Dev.} & \textbf{Mean} & \textbf{Max} & \textbf{Min} & \textbf{S.Dev.} &                                       &                                                                                                       &                                                                                                \\ \hline
\%7                                                                                                & 873.0         & 972.6        & 688.7        & 59.2            & 2229.5        & 2546.0       & 1987.2       & 140.0           & \%7                                     & 1 & 41                                                                                                    \\ \hline
\%13                                                                                               & 851.3         & 929.5        & 711.3        & 44.0            & 1596.6        & 1774.5       & 1431.6       & 80.2            & \%9                                     & 1 &41                                                                                                    \\ \hline
\%20                                                                                               & 849.1         & 924.9        & 724.4        & 40.5            & 1391.9        & 1527.4       & 1248.6       & 63.4            & \%10                                    & 1    &31                                                                                                 \\ \hline
\%26                                                                                               & 848.2         & 922.4        & 731.7        & 39.3            & 1278.3        & 1391.2       & 1148.2       & 55.0            & \%12                                    & 1 &28                                                                                                    \\ \hline
\%34                                                                                               & 847.7         & 920.9        & 737.2        & 38.7            & 1197.6        & 1293.9       & 1075.7       & 50.0            & \%14                                    & 1    &28                                                                                            \\ \hline
\%43                                                                                               & 847.3         & 921.2        & 741.8        & 38.4            & 1131.6        & 1214.2       & 1016.3       & 46.2            & \%16                                    & 1 &22                                                                                                \\ \hline
\%53                                                                                               & 847.0         & 925.1        & 745.5        & 38.3            & 1080.3        & 1154.2       & 970.4        & 43.7            & \%18                                    & 1    &16                                                                                           \\ \hline
\%64                                                                                               & 846.7         & 929.1        & 749.1        & 38.4            & 1029.0        & 1103.7       & 925.0        & 41.8            & \%22                                    & 1    &16                                                                                               \\ \hline
\%76                                                                                               & 846.4         & 932.7        & 752.4        & 38.6            & 981.3         & 1061.2       & 882.6        & 40.5            & \%28                                    & 1       &10                                                                                              \\ \hline
\%92                                                                                               & 846.1         & 937.5        & 756.8        & 38.9            & 919.0         & 1007.0       & 827.2        & 39.5            & \%47                                    & 1          &5                                                                                        \\ \hline
\end{tabular}}
\caption{Newsvendor Problem Test Data Set Statistical Values} 
\label{tab:table2}
\end{table}

\begin{figure}

\makebox[\linewidth][c]{%
\begin{subfigure}[b]{.42\textwidth}
\centering
\includegraphics[width=.95\textwidth]{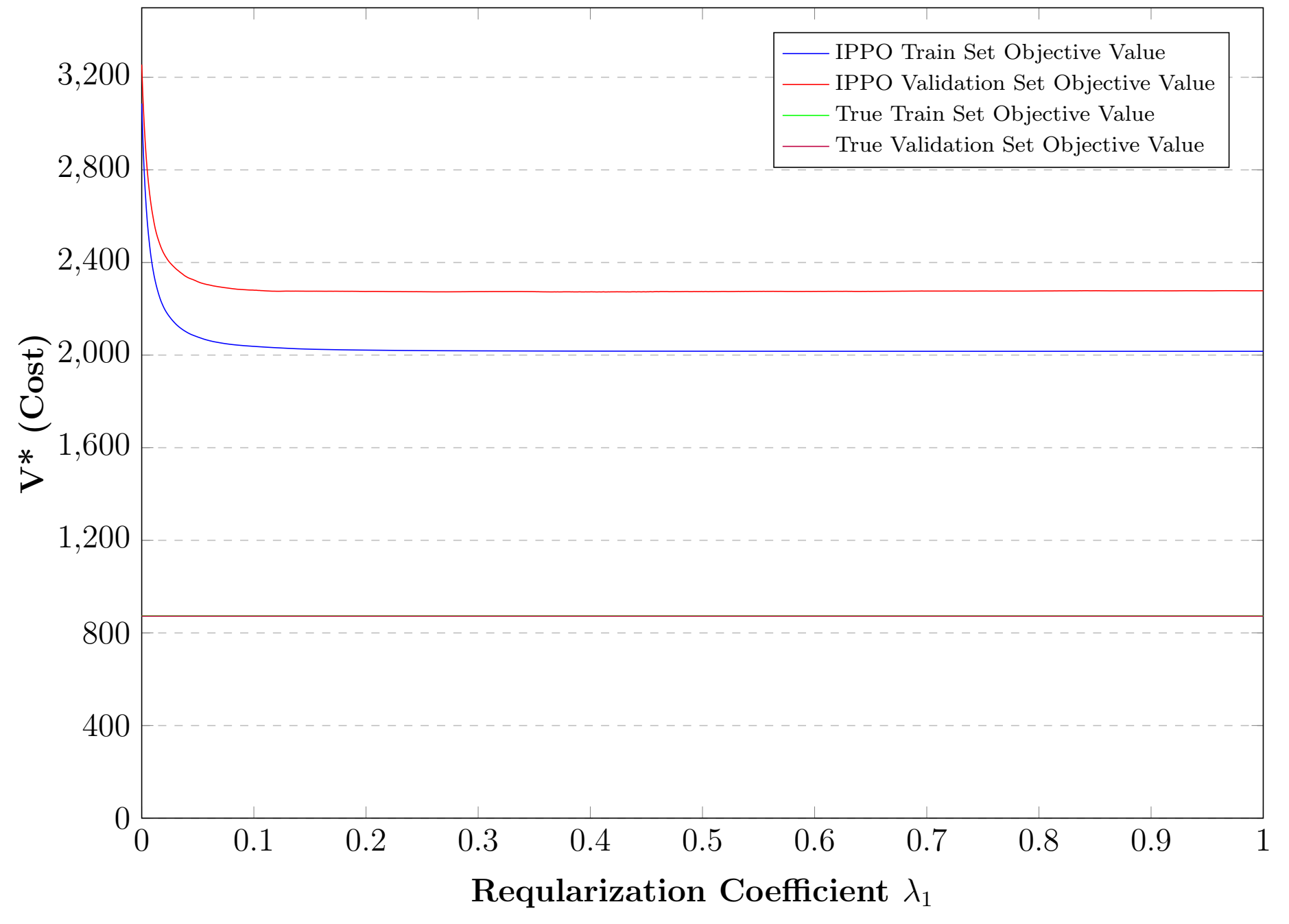}
\caption{R-Square=\%7}
\end{subfigure} \hspace{-5mm}%
\begin{subfigure}[b]{.42\textwidth}
\centering
\includegraphics[width=.95\textwidth]{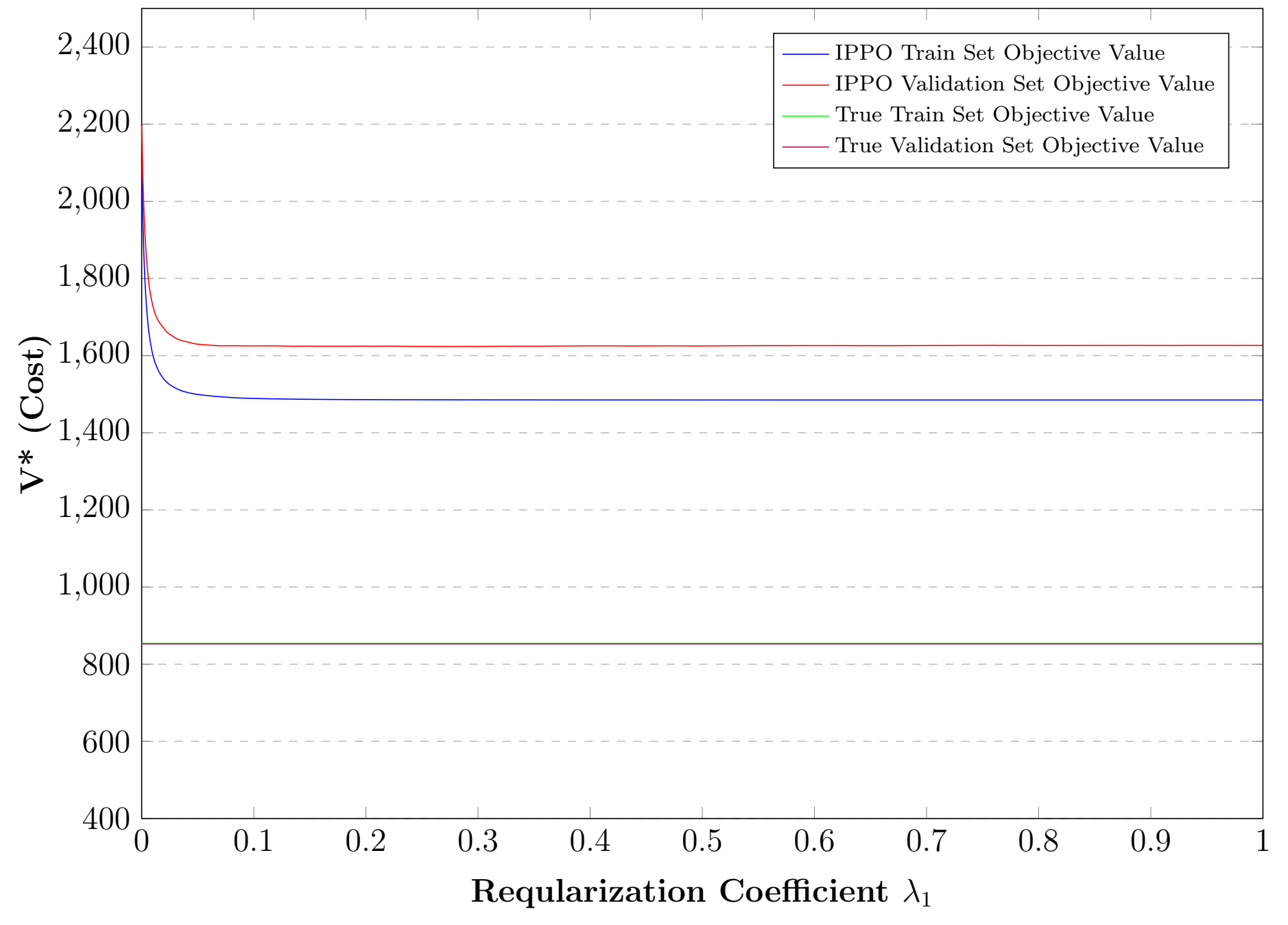}
\caption{R-Square=\%13}
\end{subfigure} \hspace{-5mm}%
\begin{subfigure}[b]{.42\textwidth}
\centering
\includegraphics[width=.95\textwidth]{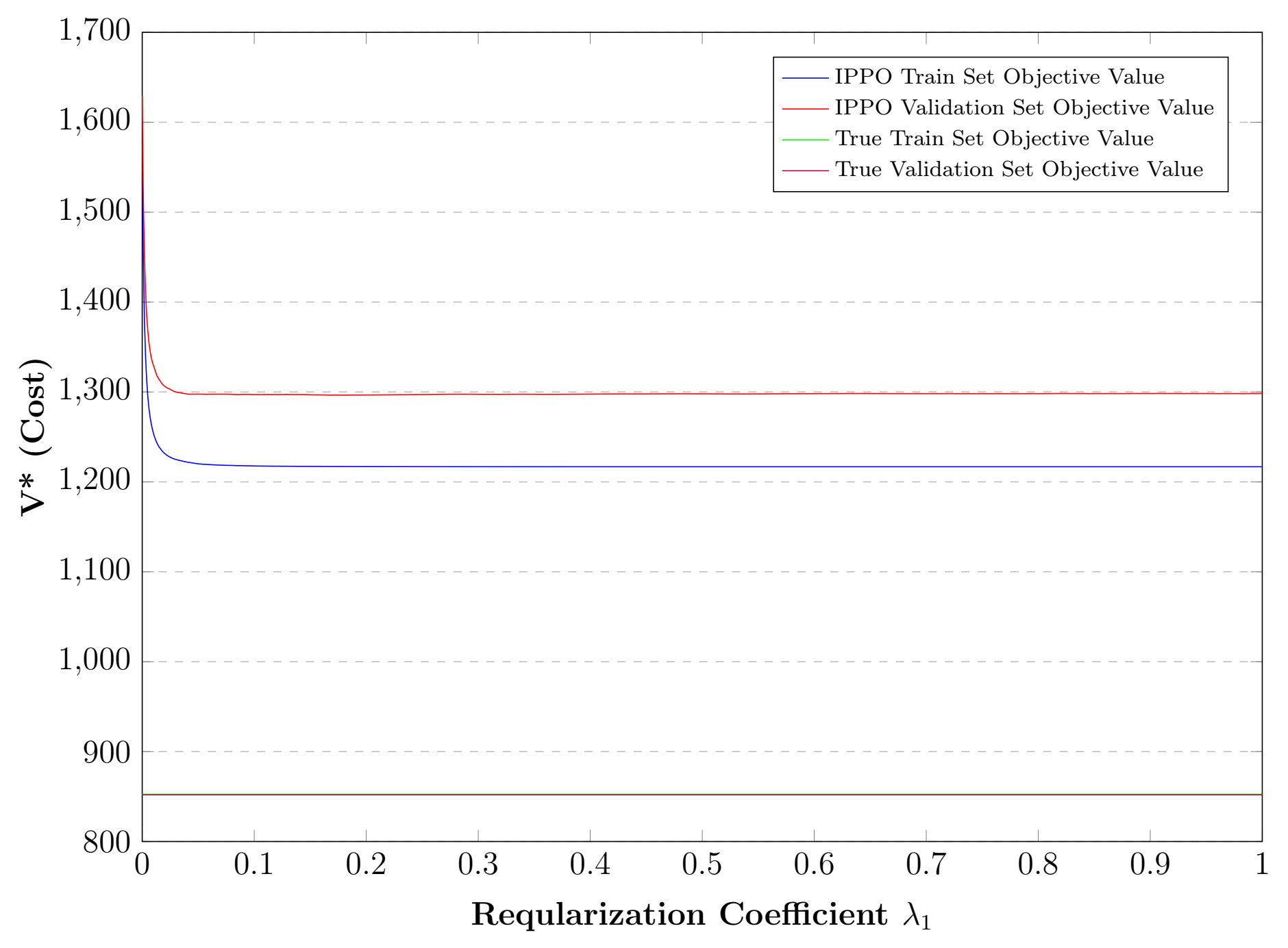}
\caption{R-Square=\%26}
\end{subfigure}%
}\\

\makebox[\linewidth][c]{%
\begin{subfigure}[b]{.42\textwidth}
\centering
\includegraphics[width=.95\textwidth]{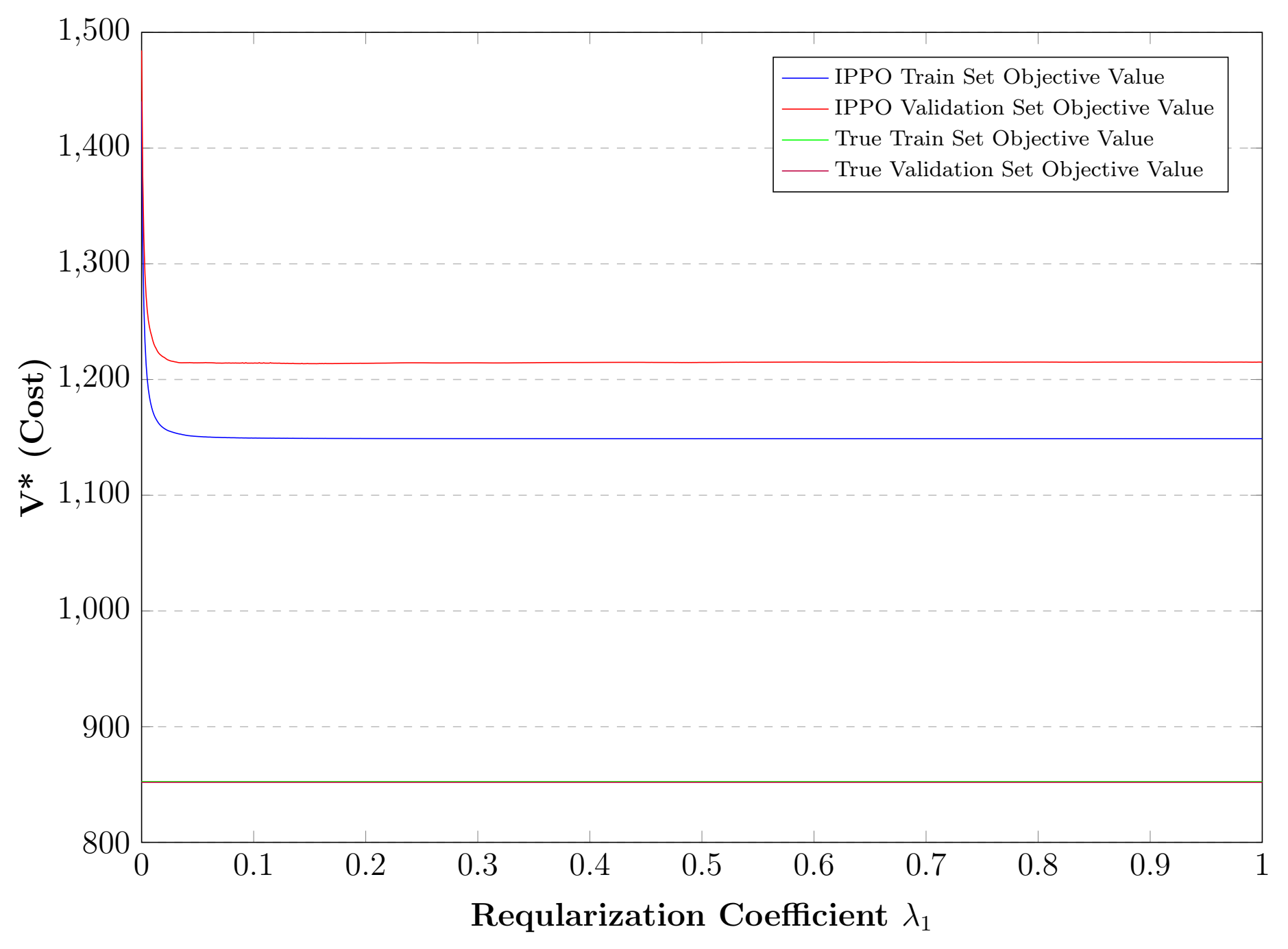}
\caption{R-Square=\%34}
\end{subfigure} \hspace{-5mm}%
\begin{subfigure}[b]{.42\textwidth}
\centering
\includegraphics[width=.95\textwidth]{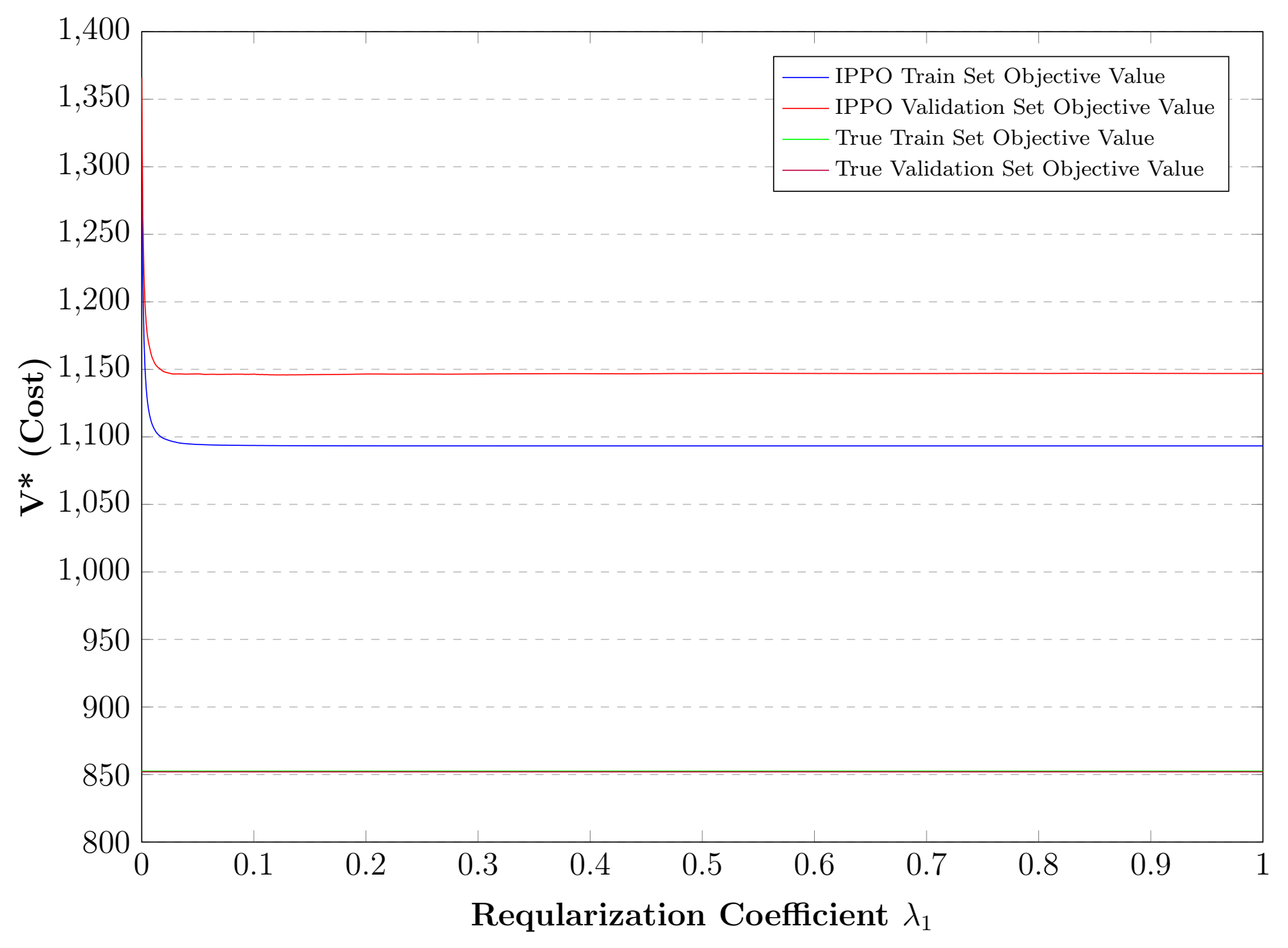}
\caption{R-Square=\%43}
\end{subfigure} \hspace{-5mm}%
\begin{subfigure}[b]{.42\textwidth}
\centering
\includegraphics[width=.95\textwidth]{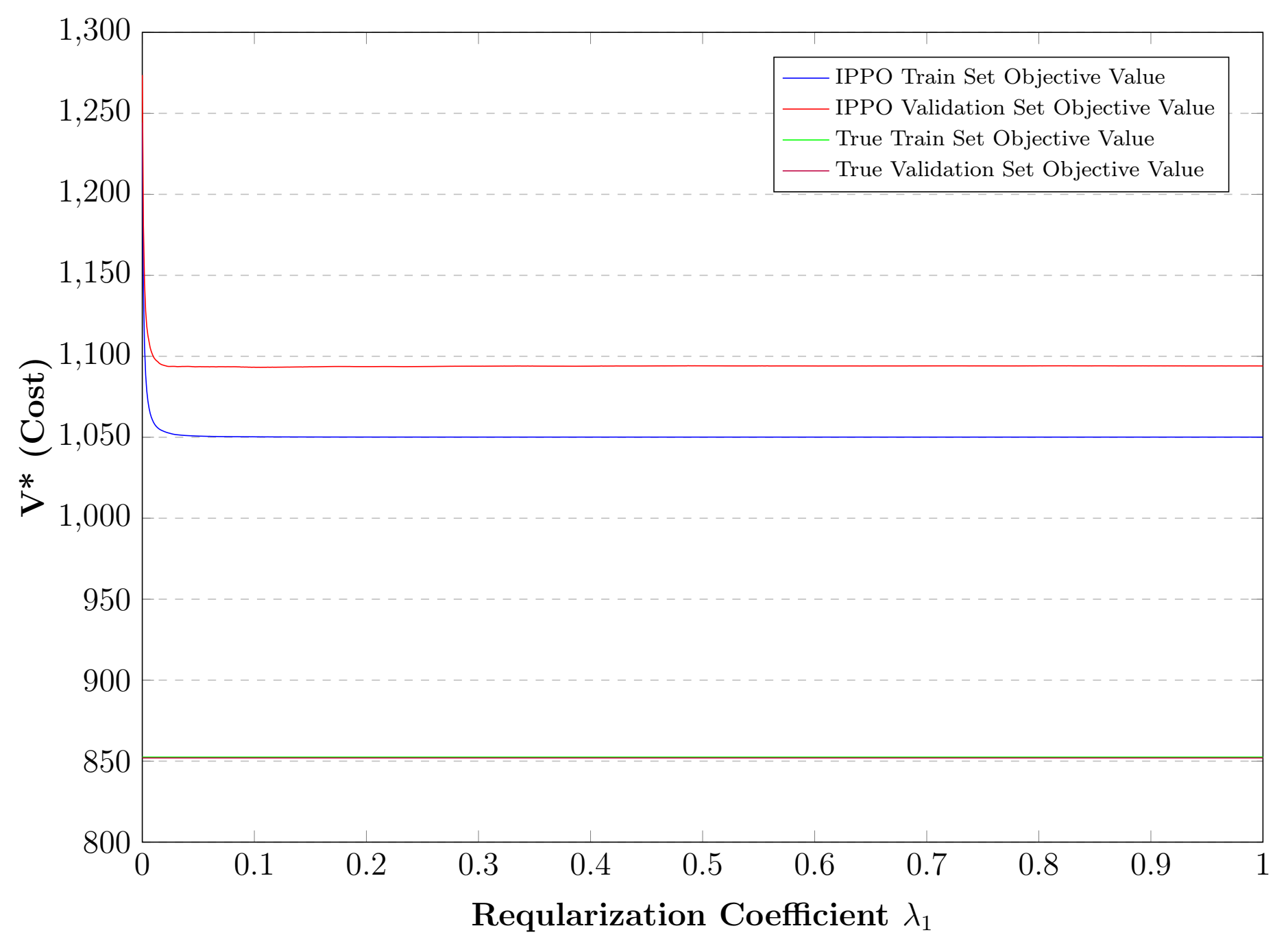}
\caption{R-Square=\%53}
\end{subfigure}%
}\\

\makebox[\linewidth][c]{%
\begin{subfigure}[b]{.42\textwidth}
\centering
\includegraphics[width=.95\textwidth]{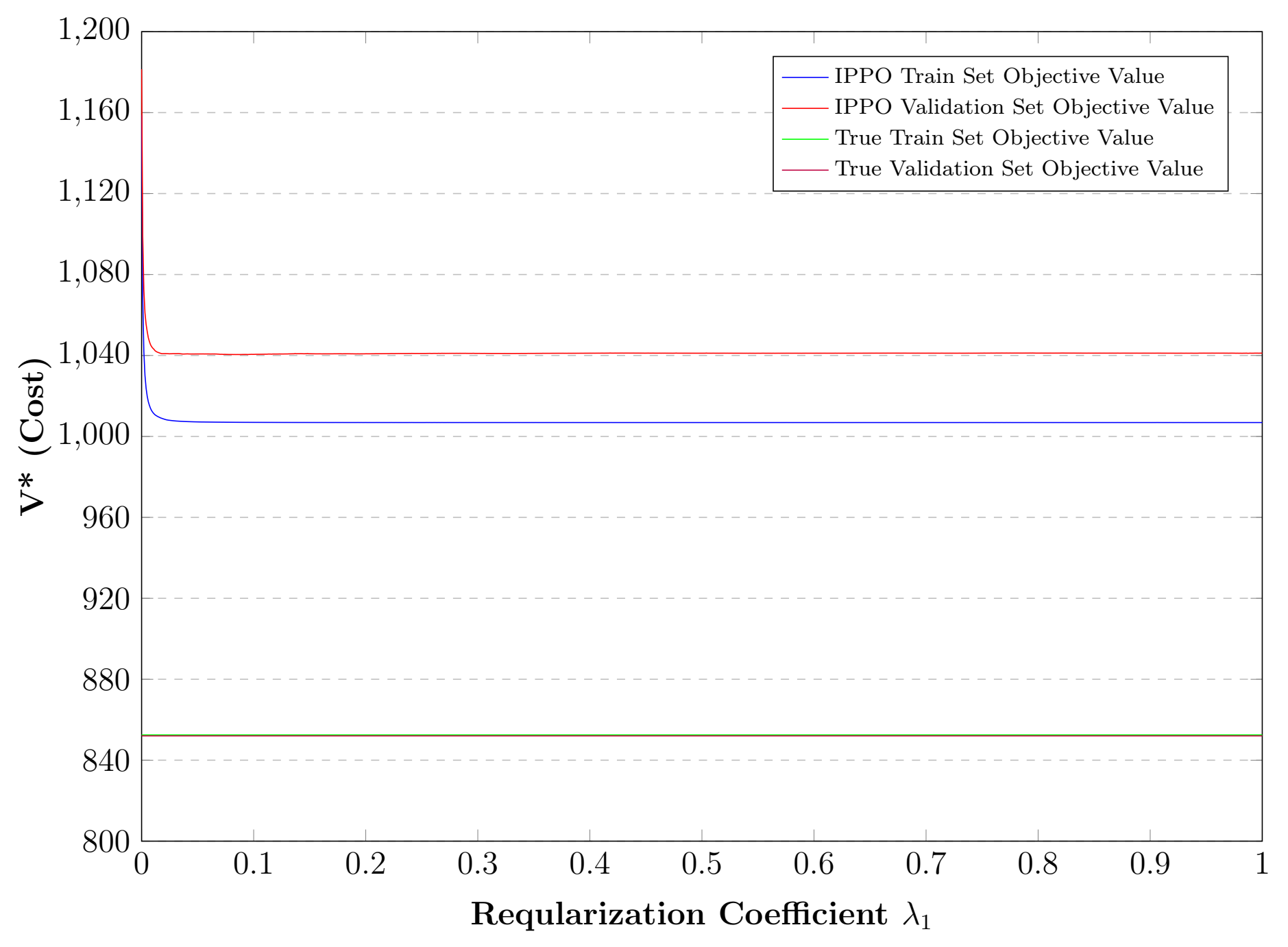}
\caption{R-Square=\%64}
\end{subfigure} \hspace{-5mm}%
\begin{subfigure}[b]{.42\textwidth}
\centering
\includegraphics[width=.95\textwidth]{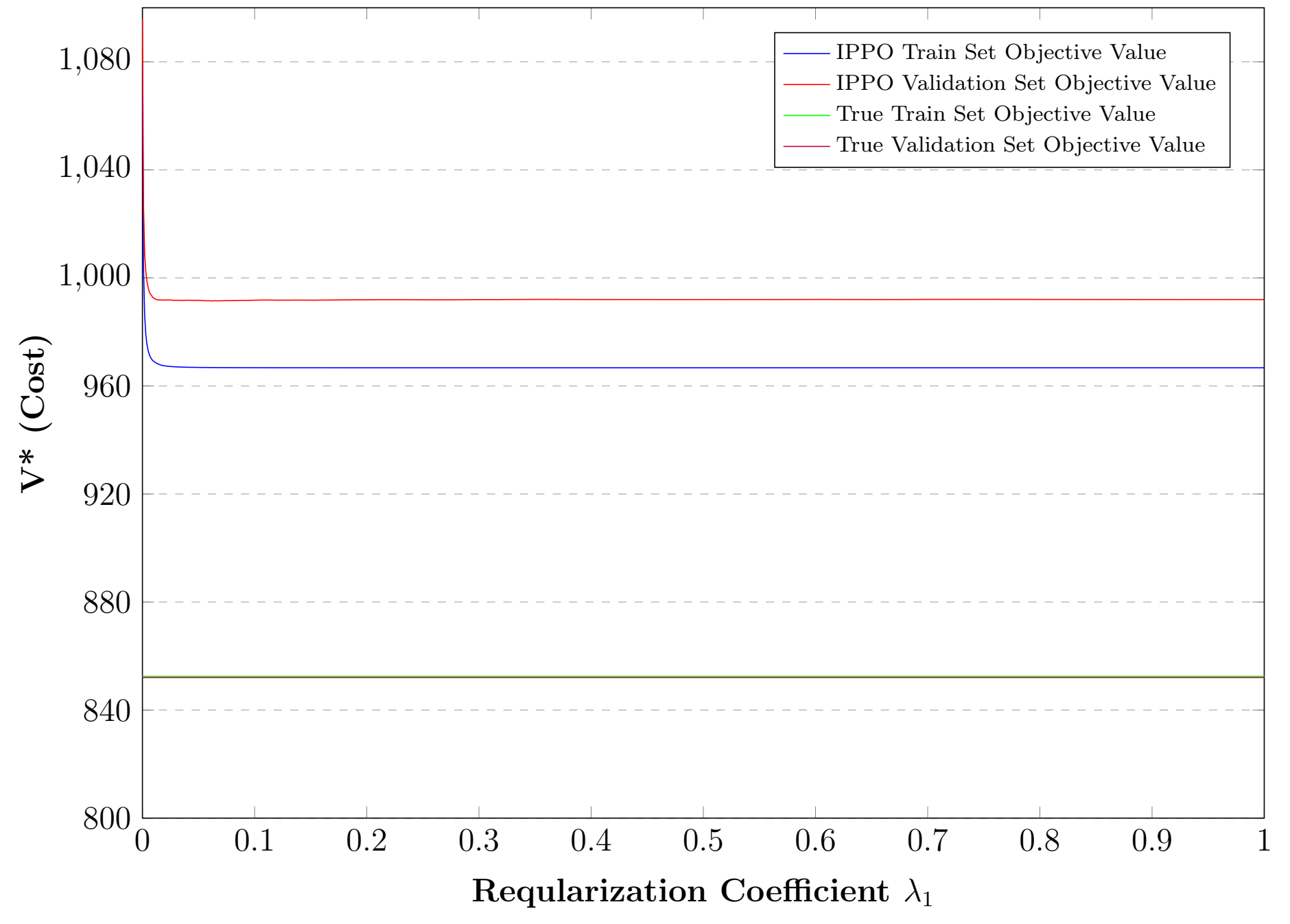}
\caption{R-Square=\%76}
\end{subfigure} \hspace{-5mm}%
\begin{subfigure}[b]{.42\textwidth}
\centering
\includegraphics[width=.95\textwidth]{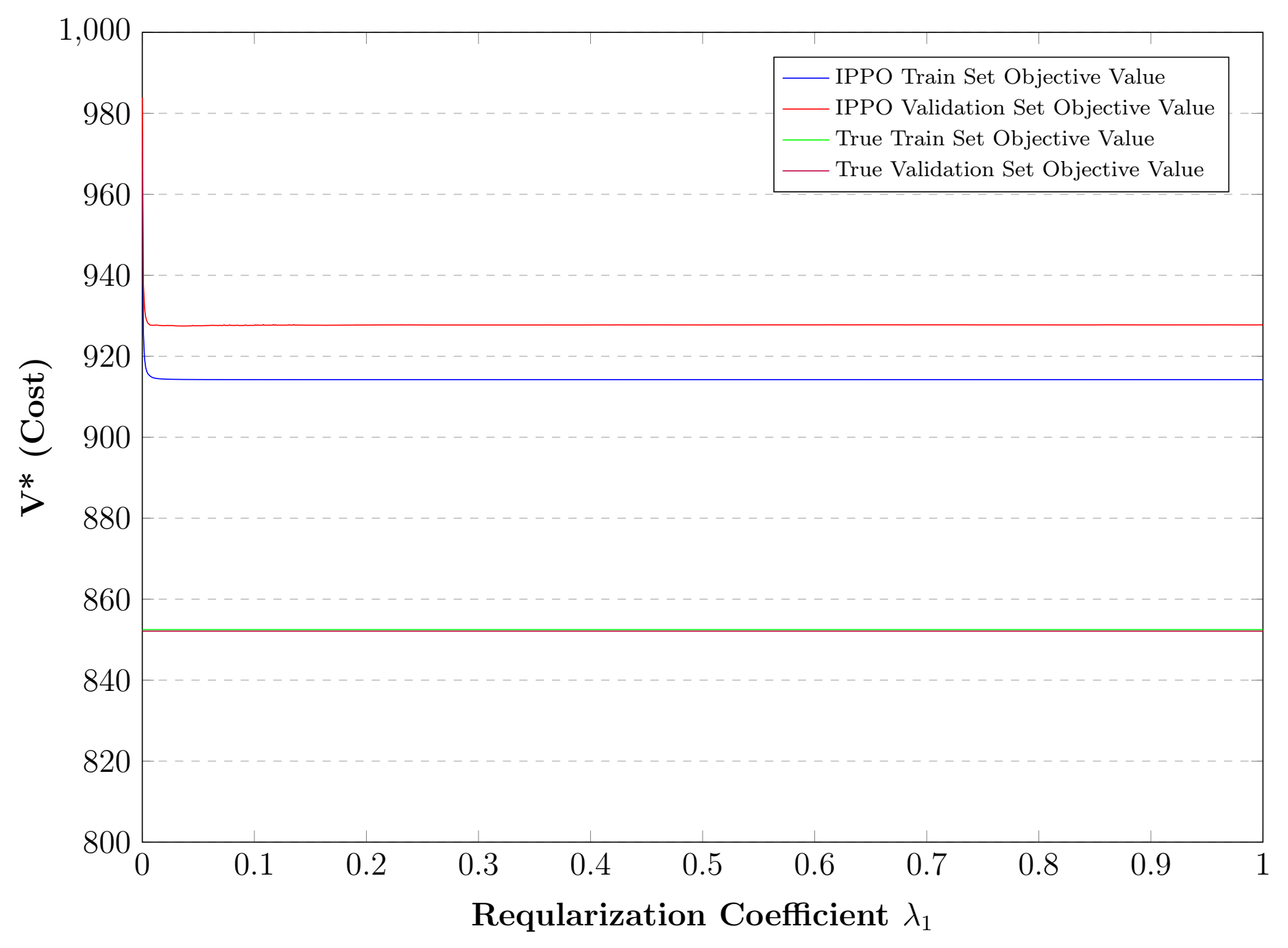}
\caption{R-Square=\%92}
\end{subfigure}%
}\\
\caption{Newsvendor Problem Train and Validation Performance Over Regularization Parameter $\lambda_1$ For Different R-Square Values Between X and Y}
\label{plot:graph4}
\end{figure}

We observe that value of k neighbors increases when correlation level between $X$ and $Y$ decreases. Since this correlation decreases, kNN tends to minimize expected cost over more train data as we reported in Table \ref{tab:table1} and \ref{tab:table2}. We did not include result of \cite{Elmachtoub2017SmartT} in Figure \ref{plot:graph3} and \ref{plot:graph5} because \cite{Elmachtoub2017SmartT} build their method when there is an uncertainty in cost vector of objective function.\\
Since IPPO methodology fully leverages side information and evaluates decisions with respect to true parameters inside of integrated framework, it gives the best solution.
In newsvendor problem, we did not observe over-fitting issue, and there is no need to control generalization error. As we report in Figure \ref{plot:graph4}, regardless of correlation level between $X$ and $Y$, validation and train performance improve constantly until $\lambda_1=1$ where no generalization is needed. That is why IPPO with and without generalization error in Figure \ref{plot:graph3} overlaps.

\subsection{Two-Stage Shipment Problem}
\begin{figure}
\makebox[\linewidth][c]{%
\begin{subfigure}[b]{.6\textwidth}
\centering
\includegraphics[width=.95\textwidth]{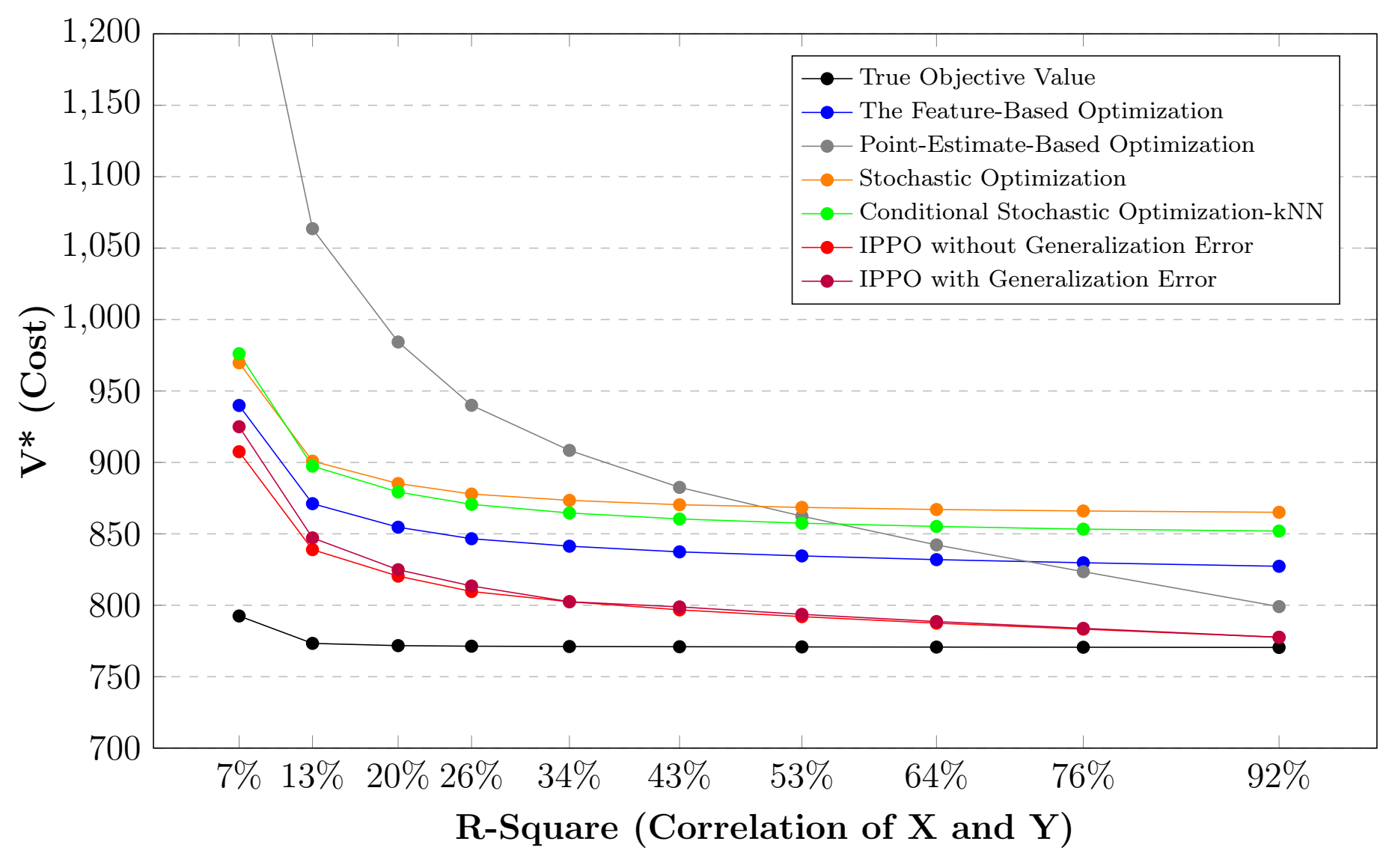}
\caption{Train Data Set}
\end{subfigure} \hspace{-5mm}%
\begin{subfigure}[b]{.6\textwidth}
\centering
\includegraphics[width=.95\textwidth]{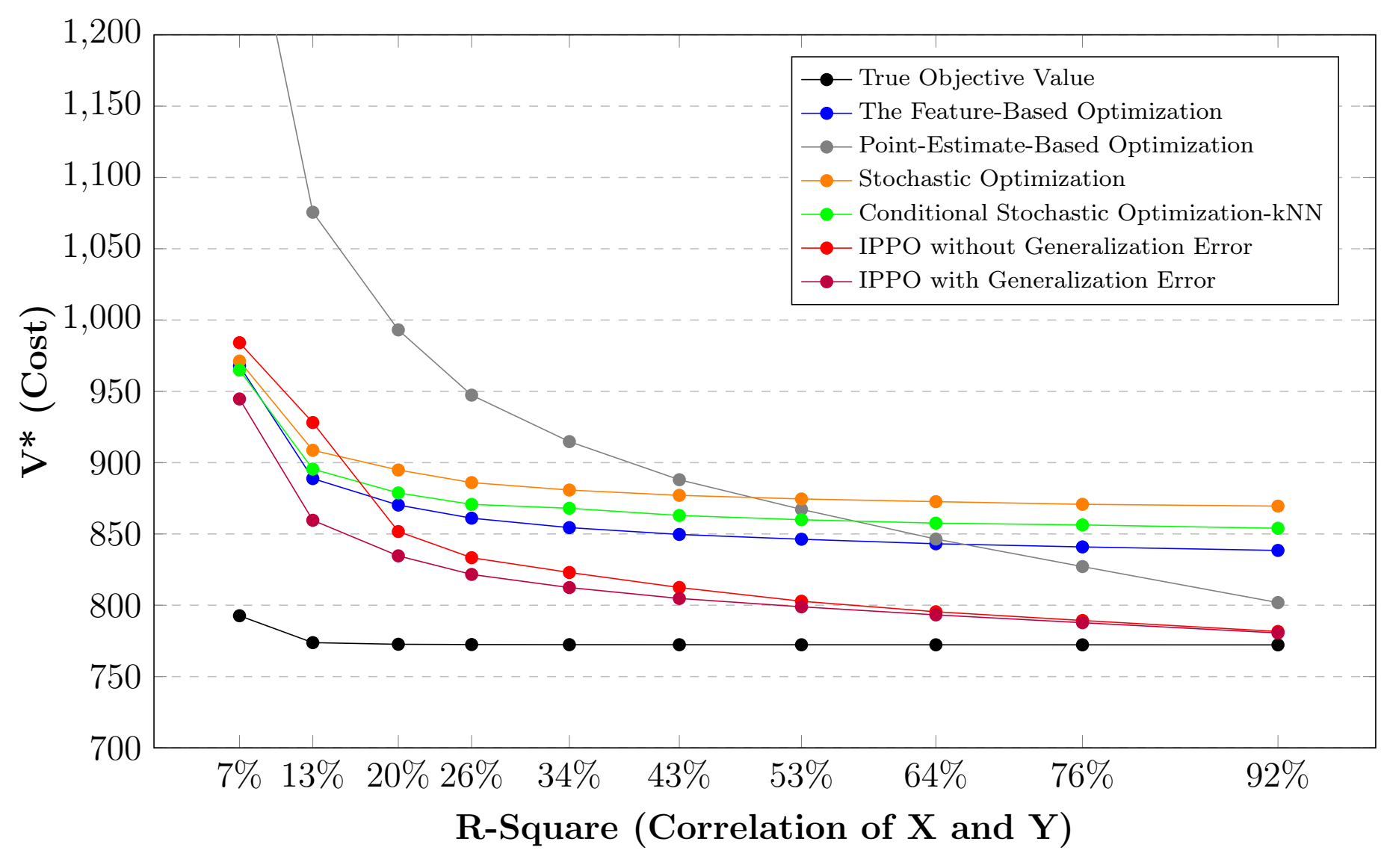}
\caption{Test Data Set}
\end{subfigure}%
}\\
\caption{Comparison of Different Methods for Shipment Problem}
\label{plot:graph5}
\end{figure}

\begin{figure}

\makebox[\linewidth][c]{%
\begin{subfigure}[b]{.42\textwidth}
\centering
\includegraphics[width=.95\textwidth]{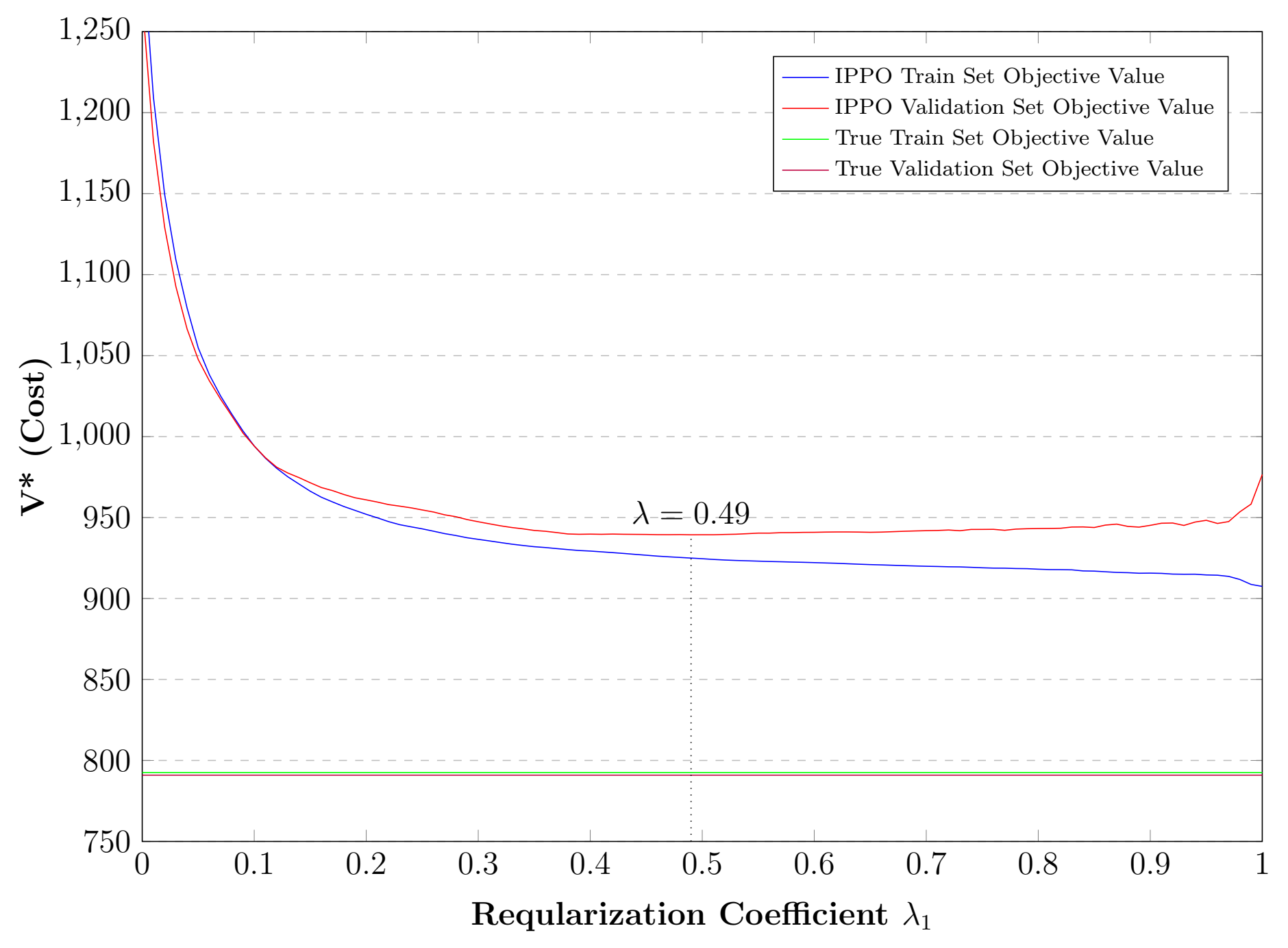}
\caption{R-Square=\%7}
\end{subfigure} \hspace{-5mm}%
\begin{subfigure}[b]{.42\textwidth}
\centering
\includegraphics[width=.95\textwidth]{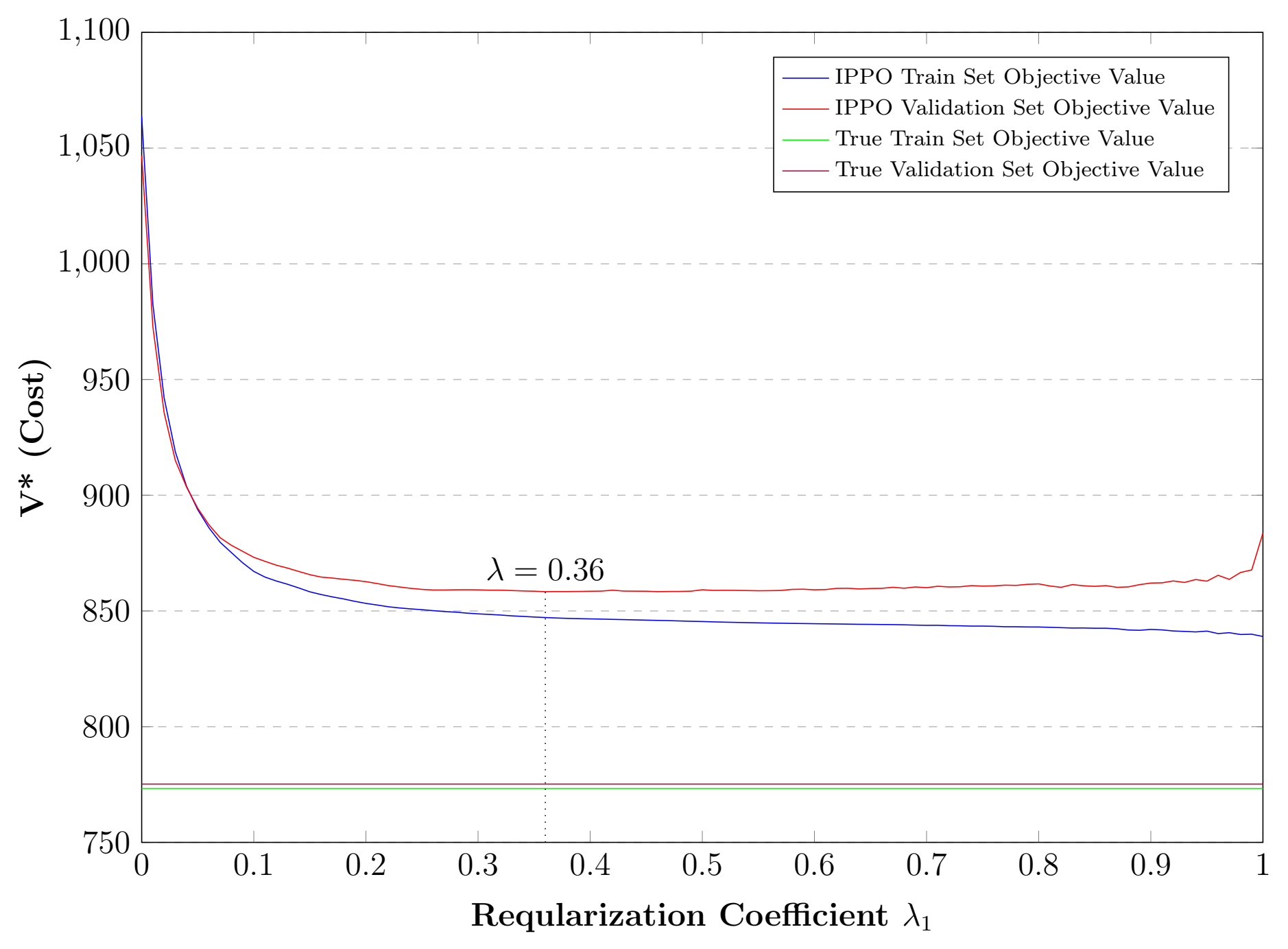}
\caption{R-Square=\%13}
\end{subfigure} \hspace{-5mm}%
\begin{subfigure}[b]{.42\textwidth}
\centering
\includegraphics[width=.95\textwidth]{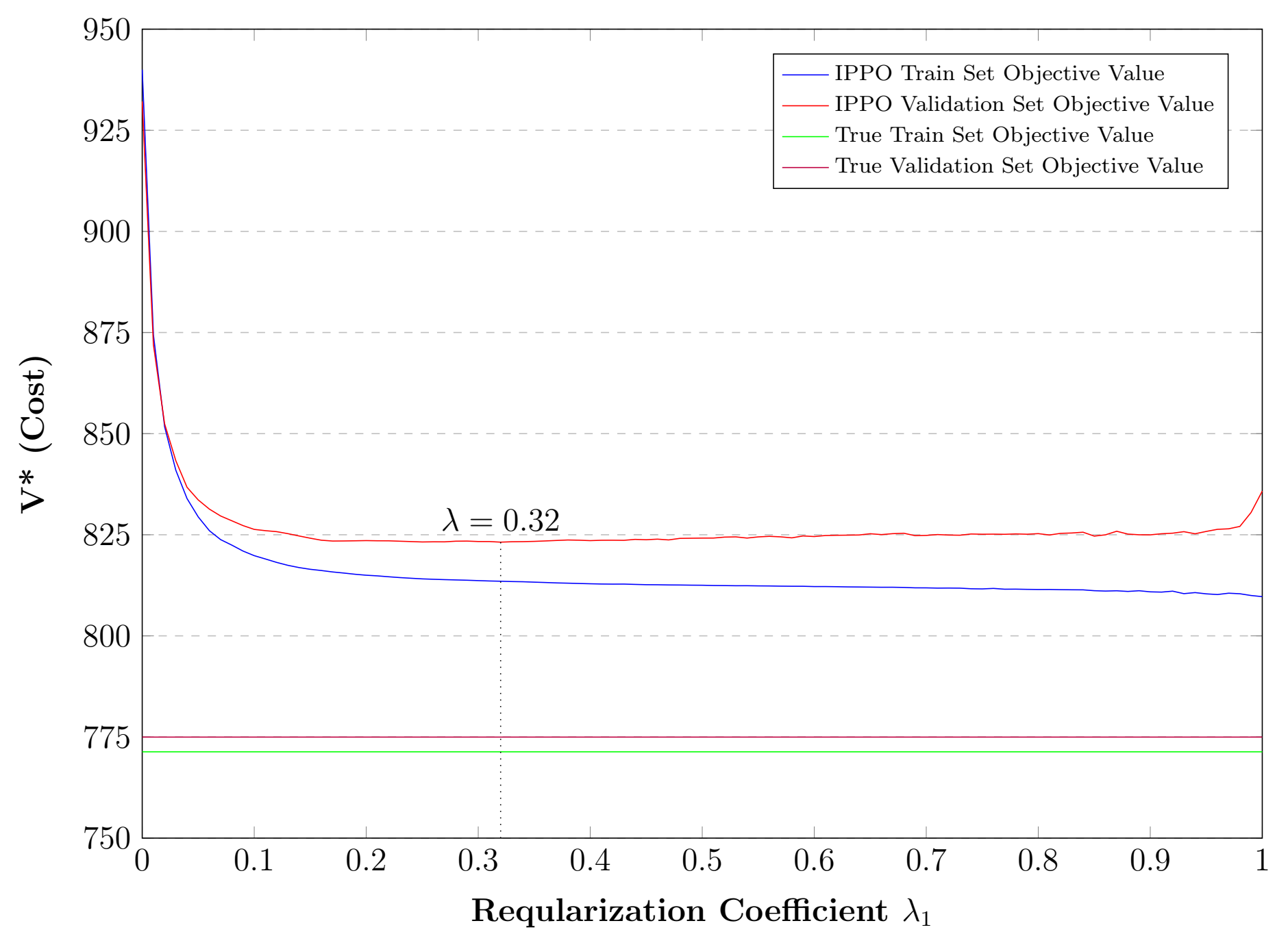}
\caption{R-Square=\%26}
\end{subfigure}%
}\\

\makebox[\linewidth][c]{%
\begin{subfigure}[b]{.42\textwidth}
\centering
\includegraphics[width=.95\textwidth]{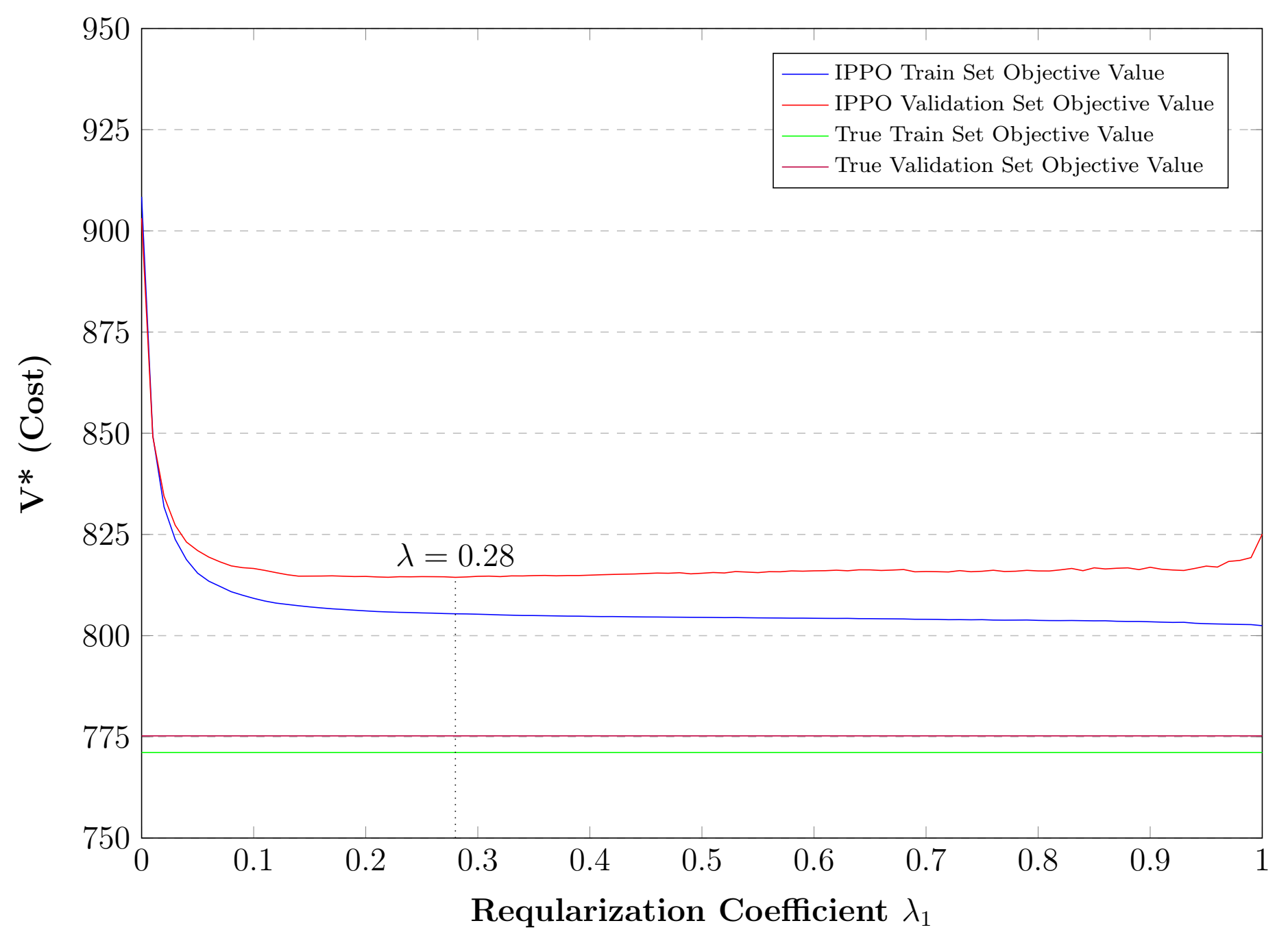}
\caption{R-Square=\%34}
\end{subfigure} \hspace{-5mm}%
\begin{subfigure}[b]{.42\textwidth}
\centering
\includegraphics[width=.95\textwidth]{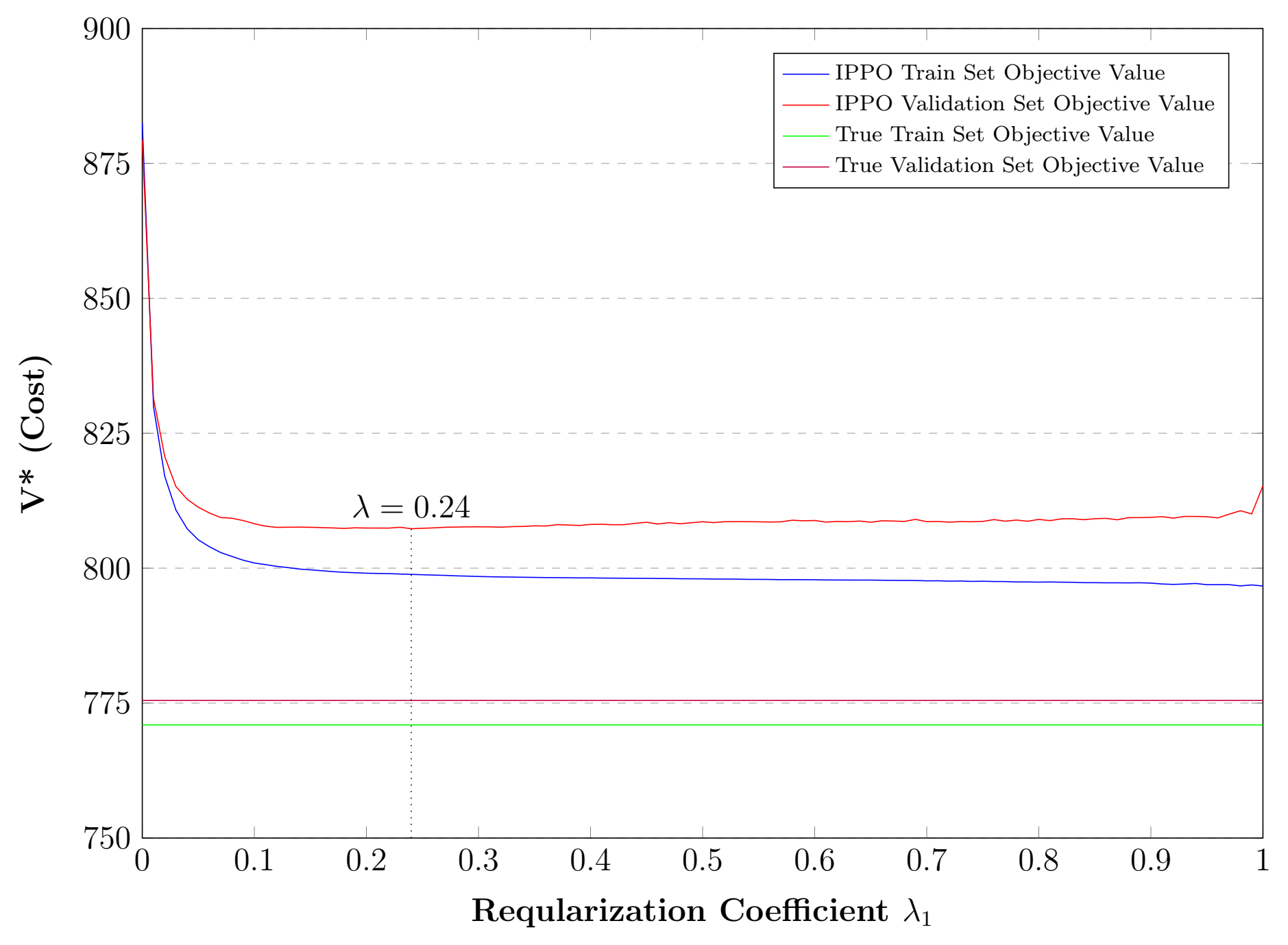}
\caption{R-Square=\%43}
\end{subfigure} \hspace{-5mm}%
\begin{subfigure}[b]{.42\textwidth}
\centering
\includegraphics[width=.95\textwidth]{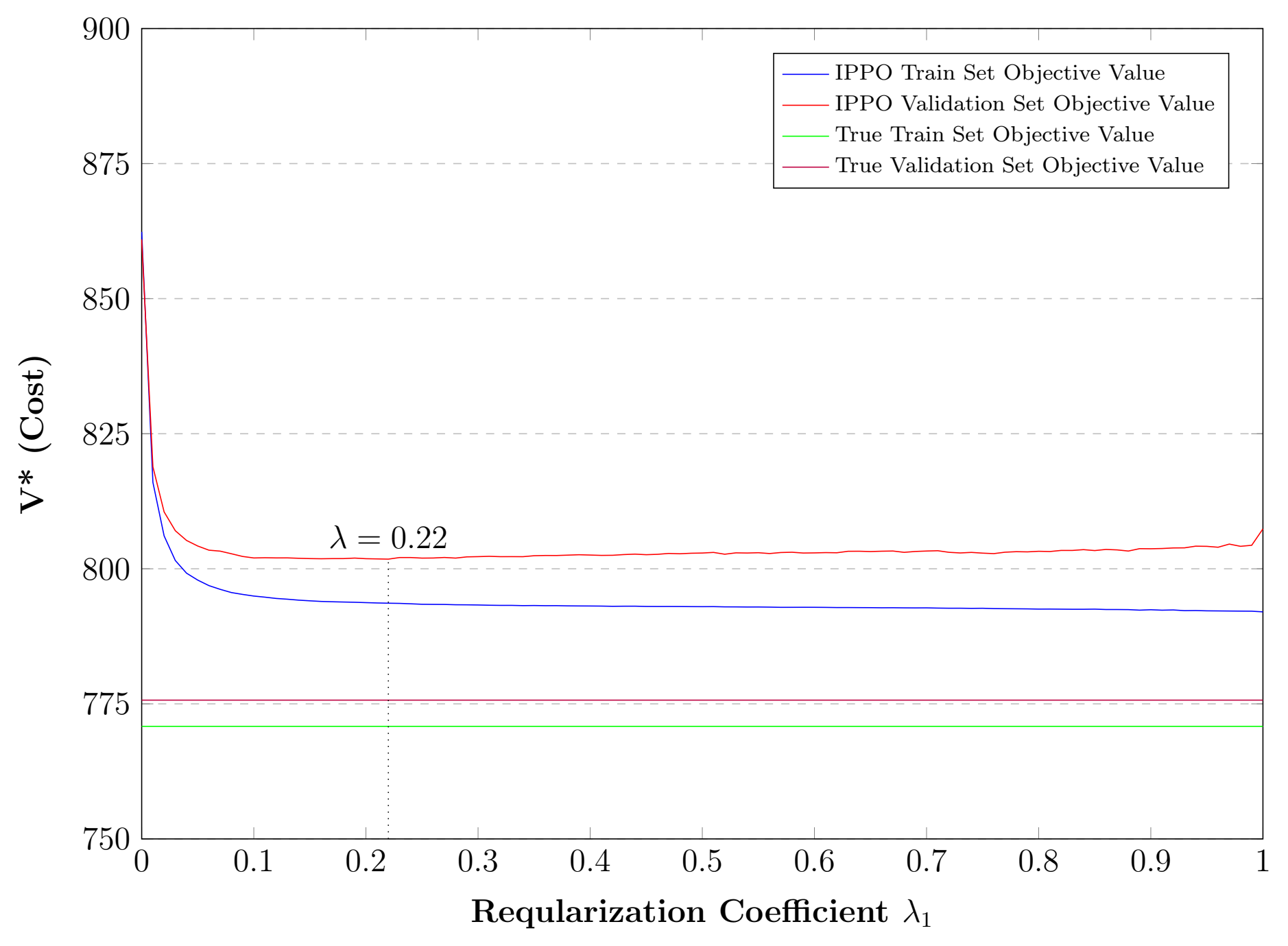}
\caption{R-Square=\%53}
\end{subfigure}%
}\\
\makebox[\linewidth][c]{%
\begin{subfigure}[b]{.42\textwidth}
\centering
\includegraphics[width=.95\textwidth]{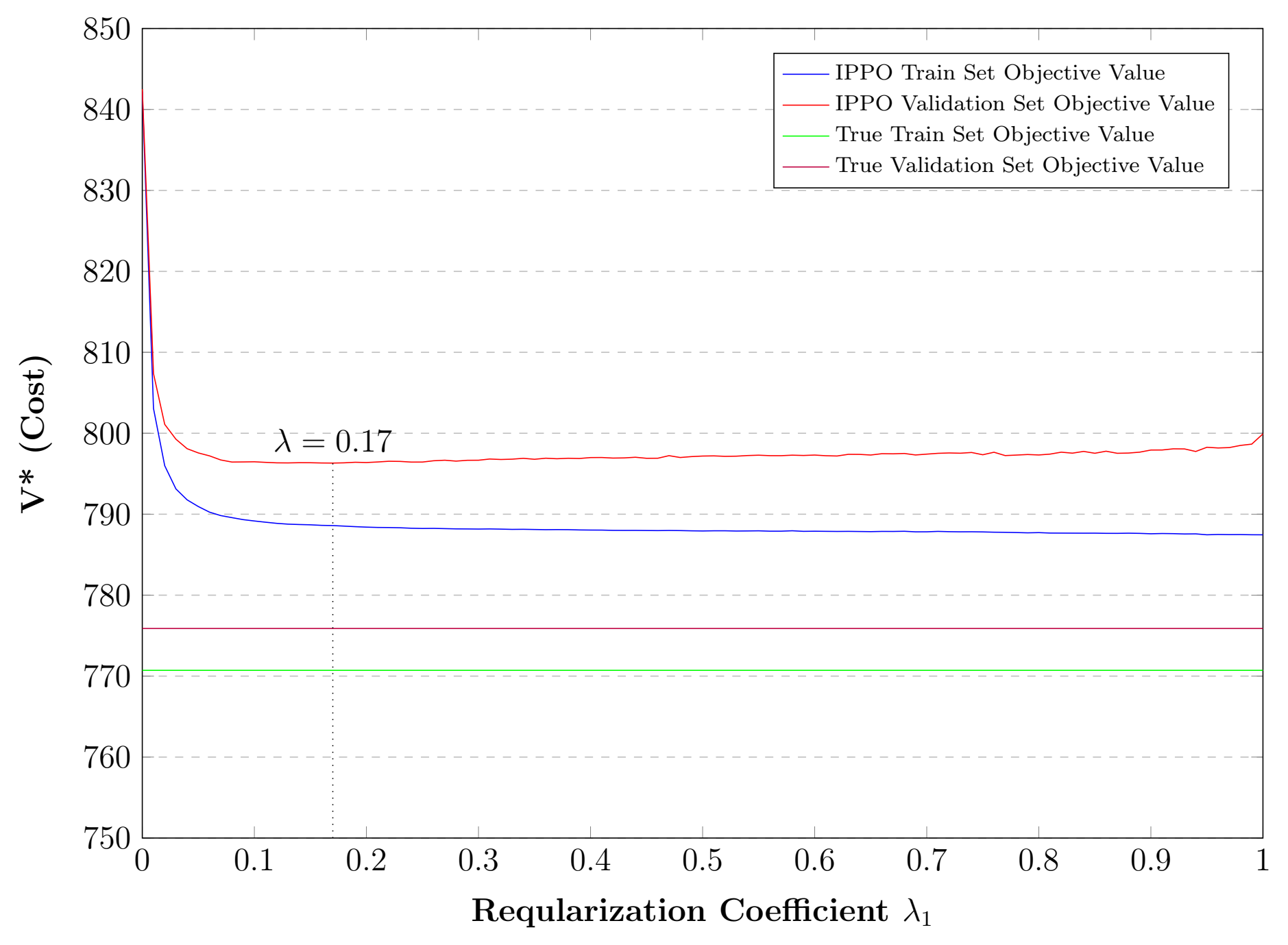}
\caption{R-Square=\%64}
\end{subfigure} \hspace{-5mm}%
\begin{subfigure}[b]{.42\textwidth}
\centering
\includegraphics[width=.95\textwidth]{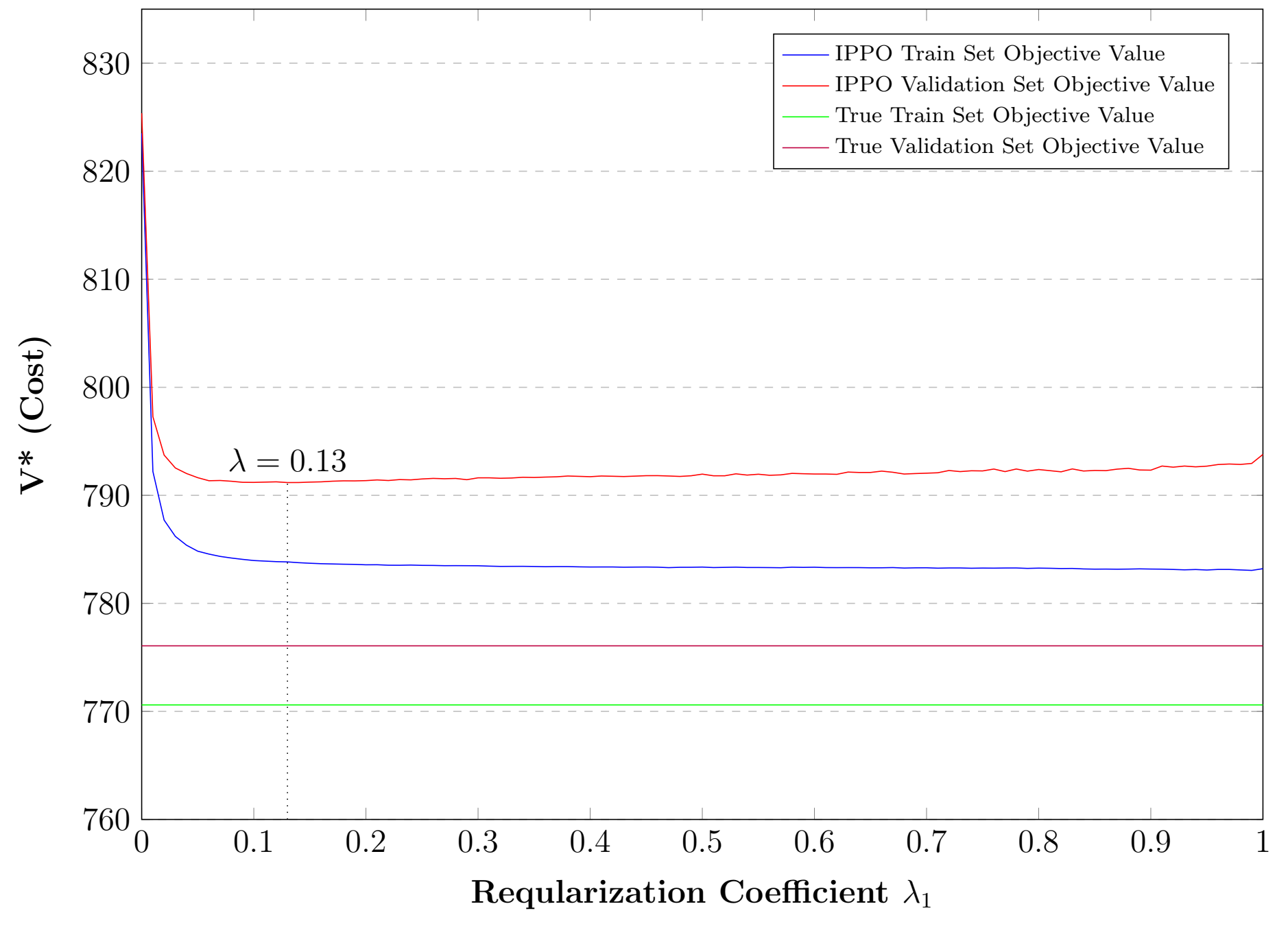}
\caption{R-Square=\%76}
\end{subfigure} \hspace{-5mm}%
\begin{subfigure}[b]{.42\textwidth}
\centering
\includegraphics[width=.95\textwidth]{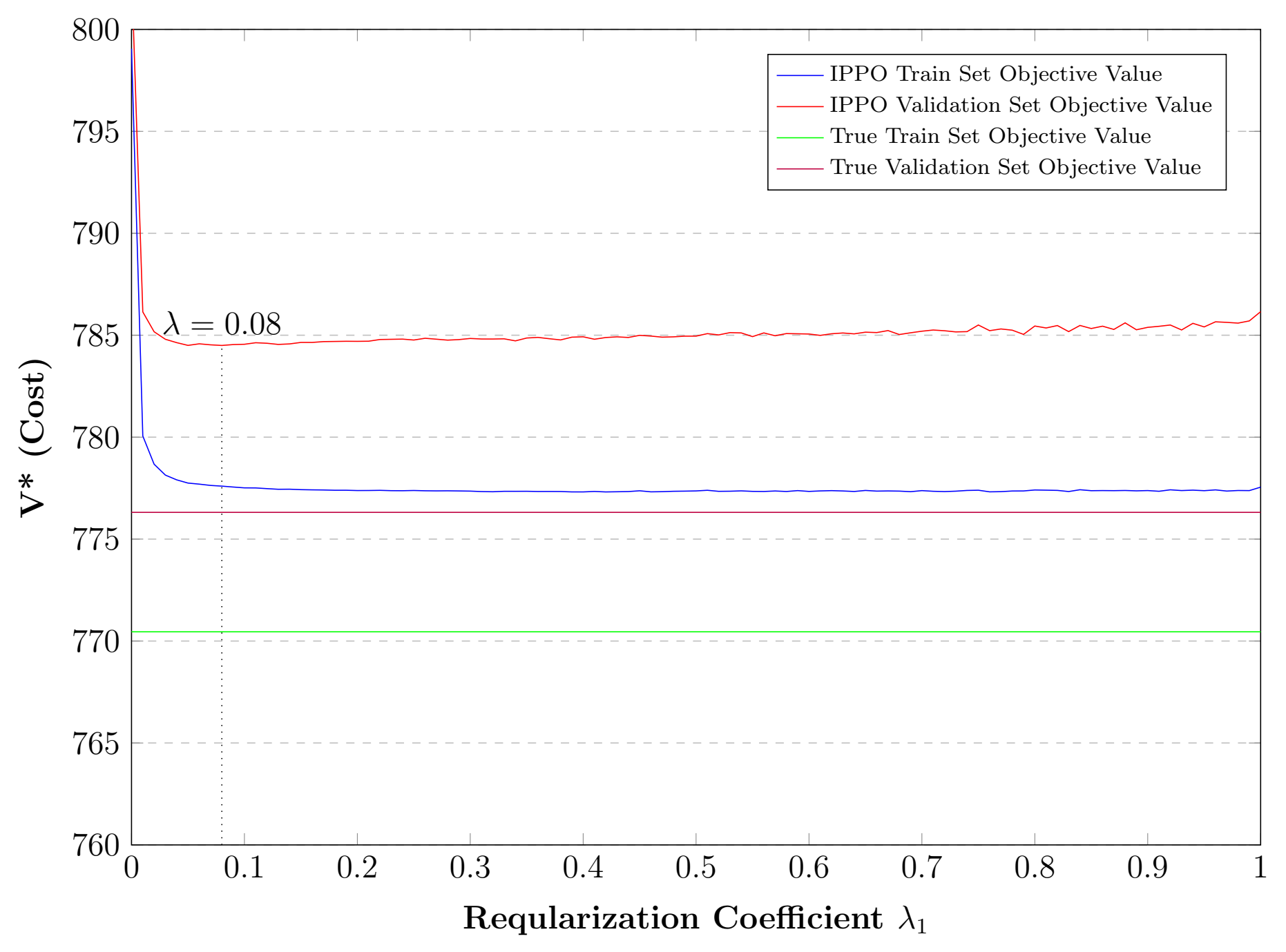}
\caption{R-Square=\%92}
\end{subfigure}%
}\\
\caption{Shipment Problem Train and Validation Performance Over Regularization Parameter $\lambda_1$ For Different R-Square Values Between X and Y}
\label{plot:graph6}

\end{figure}

We observe similar behaviors here as in newsvendor problem. Main difference in Two-Stage shipment problem is IPPO overfits in especially low level $X$ and $Y$ correlations since IPPO methodology fully leverages side information and searches best solution for train data set. As we see the effect of regularization coefficient ($\lambda_1$) in Figure \ref{plot:graph6}, train performance constantly improves, but validation performance become worse after some point. In figure \ref{plot:graph5}, we report train and test performance with and without controlling generalization error. Although train performance is better than other methods without controlling generalization error, a better solution is achieved via optimizing regularization coefficient ($\lambda_1$). Some detailed statistical measures are given in Table \ref{tab:table3} for train set and in Table \ref{tab:table4} for test set. Performance column is calculated in Equation \ref{eq:eq39} based on second best method, which is The Feature-Based Optimization. These results are based on 30 replications, mean column is presented in Figure \ref{plot:graph5}. \\[-30pt]

\begin{flalign}
Performance =\left[\frac{The \hspace{1 mm} Feature \hspace{1 mm} Based-True \hspace{1 mm} Objective}{IPPO-True \hspace{1 mm} Objective}-1\right]100 \label{eq:eq39}
\end{flalign}

\begin{table}
\resizebox{\textwidth}{!}{\begin{tabular}{|c|c|c|c|c|c|c|c|c|c|c|c|}
\hline
\multirow{2}{*}{\textbf{\begin{tabular}[c]{@{}c@{}}Correlation \\ between X and Y\end{tabular}}} & \multicolumn{4}{c|}{\textbf{True Objective Values}}           & \multicolumn{4}{c|}{\textbf{IPPO}}                            & \multirow{2}{*}{\textbf{Performance}} & \multirow{2}{*}{\textbf{\begin{tabular}[c]{@{}c@{}}Optimal Regularization \\ Parameter\end{tabular}}} & \multirow{2}{*}{\textbf{\begin{tabular}[c]{@{}c@{}}Optimal Neighbors\\ for  kNN\end{tabular}}} \\ \cline{2-9}
                                                                                                 & \textbf{Mean} & \textbf{Max} & \textbf{Min} & \textbf{S.Dev.} & \textbf{Mean} & \textbf{Max} & \textbf{Min} & \textbf{S.Dev.} &                                       &                                                                                                       &                                                                                                \\ \hline
\%7                                                                                                & 792.5         & 855.7        & 738.0        & 22.2            & 925.0         & 988.4        & 864.1        & 28.4            & \%11                                                                                          & 0.49 &50                                                                                                 \\ \hline
\%13                                                                                               & 773.3         & 820.2        & 728.2        & 18.6            & 847.2         & 889.9        & 798.1        & 21.2            & \%32                                                                                          & 0.36    &26                                                                                          \\ \hline
\%20                                                                                               & 771.7         & 812.5        & 730.8        & 17.5            & 824.9         & 862.0        & 781.6        & 18.7            & \%56                                                                                          & 0.38   &28                                                                                          \\ \hline
\%26                                                                                               & 771.3         & 808.7        & 732.7        & 17.0            & 813.5         & 847.9        & 773.0        & 17.9            & \%78                                                                                          & 0.32      &28                                                                                    \\ \hline
\%34                                                                                               & 771.1         & 806.1        & 734.0        & 16.7            & 805.4         & 838.0        & 766.8        & 17.3            & \%105                                                                                         & 0.28  &17                                                                                                \\ \hline
\%43                                                                                               & 771.0         & 804.0        & 735.0        & 16.5            & 798.9         & 830.0        & 761.6        & 16.9            & \%138                                                                                         & 0.24     &15                                                                                             \\ \hline
\%53                                                                                               & 770.8         & 802.4        & 735.8        & 16.4            & 793.6         & 823.7        & 757.7        & 16.7            & \%179                                                                                         & 0.22        &15                                                                                          \\ \hline
\%64                                                                                               & 770.7         & 801.0        & 736.7        & 16.3            & 788.6         & 818.7        & 753.7        & 16.5            & \%242                                                                                         & 0.17           &15                                                                                    \\ \hline
\%76                                                                                               & 770.6         & 801.7        & 737.4        & 16.2            & 783.8         & 814.9        & 750.1        & 16.3            & \%347                                                                                         & 0.13              &15                                                                                    \\ \hline
\%92                                                                                               & 770.5         & 802.7        & 737.2        & 16.1            & 777.6         & 809.8        & 743.6        & 16.1            & \%695                                                                                         & 0.08                 &13                                                                                 \\ \hline
\end{tabular}}
\caption{Shipment Problem Train Data Set Statistical Values between True Objective and IPPO Objective} 
\label{tab:table3}
\end{table} 
\begin{table}
\resizebox{\textwidth}{!}{\begin{tabular}{|c|c|c|c|c|c|c|c|c|c|c|c|}
\hline
\multirow{2}{*}{\textbf{\begin{tabular}[c]{@{}c@{}}Correlation \\ between X and Y\end{tabular}}} & \multicolumn{4}{c|}{\textbf{True Objective Values}}           & \multicolumn{4}{c|}{\textbf{IPPO}}                            & \multirow{2}{*}{\textbf{Performance}} & \multirow{2}{*}{\textbf{\begin{tabular}[c]{@{}c@{}}Optimal Regularization \\ Parameter\end{tabular}}} & \multirow{2}{*}{\textbf{\begin{tabular}[c]{@{}c@{}}Optimal Neighbors\\ for  kNN\end{tabular}}} \\ \cline{2-9}
                                                                                                 & \textbf{Mean} & \textbf{Max} & \textbf{Min} & \textbf{S.Dev.} & \textbf{Mean} & \textbf{Max} & \textbf{Min} & \textbf{S.Dev.} &                                       &                                                                                                       &                                                                                                \\ \hline
\%7                                                                                                & 792.6         & 903.0        & 691.6        & 51.3            & 944.6         & 1089.4       & 821.4        & 67.6            & \%15                                    & 0.49 &50                                                                                                 \\ \hline
\%13                                                                                               & 773.8         & 857.3        & 690.3        & 40.0            & 859.6         & 932.8        & 766.6        & 46.8            & \%34                                    & 0.36    &26                                                                                              \\ \hline
\%20                                                                                               & 772.6         & 845.3        & 695.1        & 36.3            & 834.6         & 901.8        & 751.0        & 40.0            & \%57                                    & 0.38       &28                                                                                           \\ \hline
\%26                                                                                               & 772.4         & 838.6        & 697.9        & 34.5            & 821.6         & 886.0        & 742.4        & 37.2            & \%80                                    & 0.32          &28                                                                                        \\ \hline
\%34                                                                                               & 772.4         & 834.8        & 700.3        & 33.4            & 812.4         & 874.8        & 736.5        & 35.4            & \%105                                   & 0.28             &17                                                                                     \\ \hline
\%43                                                                                               & 772.3         & 833.2        & 702.3        & 32.5            & 804.8         & 865.6        & 731.1        & 34.0            & \%138                                   & 0.24                &15                                                                                  \\ \hline
\%53                                                                                               & 772.3         & 831.9        & 703.6        & 31.9            & 798.9         & 858.5        & 728.0        & 33.0            & \%178                                   & 0.22                   &15                                                                               \\ \hline
\%64                                                                                               & 772.3         & 830.6        & 704.1        & 31.4            & 793.2         & 851.4        & 724.3        & 32.1            & \%238                                   & 0.17                      &15                                                                            \\ \hline
\%76                                                                                               & 772.2         & 829.4        & 704.6        & 30.9            & 787.7         & 844.8        & 720.9        & 31.4            & \%343                                   & 0.13                         &15                                                                         \\ \hline
\%92                                                                                               & 772.2         & 827.8        & 705.3        & 30.4            & 780.6         & 836.1        & 714.4        & 30.6            & \%691                                   & 0.08                        &13                                                                          \\ \hline\end{tabular}}
\caption{Shipment Problem Test Data Set Statistical Values between True Objective and IPPO Objective} 
\label{tab:table4}
\end{table}

\section{Conclusion}
In this work, we provide an integrated framework fully leveraging feature data to predict responds in order to take best actions in prescriptive task. This methodology can be employed in any prescriptive task whose parameters uncertain with feature data as long as prescriptive model is convex because KKT optimality conditions requires duality, hence limited to convexity. To be able optimize predictive and prescriptive tasks at the same time, predictive task must be in a linear form, so this limits the complexity of prediction task. However, alternative methods can be used to capture nonlinearity outside of proposed framework. Our purpose is to train predictive model based on not predictive error, but also based on prescriptive error which is the assessment of decision variables provided by predicted responses with respect to true responses. Bilevel models are NP-hard problems, and it is not easy to solve, but our framework is a special case, where all decision variables are based on scenarios except the parameters of predictive model. Thus this formulation can be easily decomposed by decoupling the parameters of predictive model, and solved by PHA step by step. Also, we demonstrate the behavior of our and other frameworks under different correlation of feature and response data. Finally, we provide our results and compare them to traditional and recently introduced methods, and we perform well with and even without controlling generalization error.
.\\

\clearpage

\addcontentsline{toc}{section}{References}

\bibliography{bibliography.bib}

\bibliographystyle{abbrv}
\clearpage
\end{document}